%% file: main.tex
\documentclass[11pt]{article}

\usepackage[final]{acl}

\usepackage{times}
\usepackage{latexsym}

\usepackage[T1]{fontenc}

\usepackage[utf8]{inputenc}

\usepackage{microtype}

\usepackage{inconsolata}

\usepackage{graphicx}
\usepackage{enumitem}
\usepackage{amssymb}
\usepackage{amsmath}

\usepackage{makecell}
\usepackage{booktabs} 
\usepackage{multirow}
\usepackage{array} 
\usepackage{footnote}      
\makesavenoteenv{table}   
\usepackage{color}
\usepackage{tcolorbox}
\usepackage{soul}
\tcbuselibrary{skins}
\tcbuselibrary{breakable}
\usepackage{placeins} 

\usepackage[textsize=tiny]{todonotes}
\usepackage[fixed]{fontawesome5}
\usepackage{multirow}
\usepackage{multicol}
\usepackage{caption}
\usepackage{mathrsfs}
\usepackage{cuted}
\usepackage{float}
\usepackage{hyperref}

\usepackage{dblfloatfix}

\usepackage{colortbl}
\definecolor{my_green}{RGB}{51,102,0}
\definecolor{my_red}{RGB}{204, 0, 0}
\definecolor{my_purple}{RGB}{160, 43, 147}
\definecolor{my_blue}{RGB}{15, 158, 213}
\definecolor{my_orange}{RGB}{255, 127, 0}
\definecolor{my_brown}{RGB}{139, 69, 19}
\definecolor{my_teal}{RGB}{0, 128, 128} 
\newcommand{\ours}{{STEP-HRL}}

\title{Hierarchical Reinforcement Learning with Augmented Step-Level Transitions for LLM Agents}

\author{
 \textbf{Shuai Zhen\textsuperscript{1}},
 \textbf{Yanhua Yu\textsuperscript{1*}},
 \textbf{Roupei Guo\textsuperscript{2*}},
 \textbf{Nan Cheng\textsuperscript{2}},
 \textbf{Yang Deng\textsuperscript{3}}
\\
\\
 \textsuperscript{1}Beijing University of Posts and Telecommunications
\\
 \textsuperscript{2}China Mobile Group Design Institute Co., Ltd
\\
 \textsuperscript{3}Singapore Management University
\\
 \small{
   \textsuperscript{*}\textbf{Correspondence:} \href{mailto:yuyanhua@bupt.edu.cn}{yuyanhua@bupt.edu.cn}, \href{mailto:guoruopei@cmdi.chinamobile.com}{guoruopei@cmdi.chinamobile.com}
 }
}

\begin{document}
\maketitle
\begin{abstract}
Large language model (LLM) agents have demonstrated strong capabilities in complex interactive decision-making tasks.
However, existing LLM agents typically rely on increasingly long interaction histories, resulting in high computational cost and limited scalability.
In this paper, we propose \textbf{STEP-HRL}, a hierarchical reinforcement learning (HRL) framework that enables step-level learning by conditioning only on single-step transitions rather than full interaction histories.
STEP-HRL structures tasks hierarchically, using completed subtasks to represent \emph{global progress} of overall task. By introducing a \emph{local progress} module, it also iteratively and selectively summarizes interaction history within each subtask to produce a compact summary of local progress.
Together, these components yield augmented step-level transitions for both high-level and low-level policies.
Experimental results on ScienceWorld and ALFWorld benchmarks consistently demonstrate that STEP-HRL substantially outperforms baselines in terms of performance and generalization while reducing token usage. Our code is available at \url{https://github.com/TonyStark042/STEP-HRL}.
\end{abstract}

\section{Introduction}
Large language models (LLMs) have demonstrated remarkable capabilities as autonomous agents in sequential decision-making tasks, exhibiting sophisticated reasoning and planning abilities across diverse interactive environments~\citep{wangvoyager,yao2022react,li2022pre,ren2025r2dqg}.
To further enhance the effectiveness of autonomous agents, reinforcement learning (RL) offers a principled mechanism for enhancing agent decision-making capabilities ~\citep{xu2024strategicAgents,pang2024kalm,peiyuan2024agile}. Unlike  supervised approaches that rely solely on fixed demonstrations~\citep{zeng-etal-2024-agenttuning,lin2023swiftsage}, RL enables agents to refine policy through environmental interaction and reward feedback, thereby discovering more effective strategies that generalize beyond training distributions.

Despite this progress, most LLM agents adopt a \emph{history-conditioned} formulation, where policies are conditioned on increasingly long sequences of past observations and actions. This design choice is largely inherited from sequence-modeling perspectives: LLM agents are built on Transformer architectures~\citep{vaswani2017attention}, and recent RL formulations cast decision-making as trajectory or sequence prediction~\citep{chen2021decision,janner2021offline,ni2023transformers}. While long histories can help infer latent states in partially observable environments, conflating long-horizon decision-making with long-context conditioning introduces fundamental limitations. Attention-based inference scales quadratically with context length, and unfiltered histories accumulate redundant or irrelevant information that can obscure decision-critical signals and degrade reasoning quality~\citep{zhou2025mem1,cherepanov2023recurrent}. Importantly, long-context conditioning is a modeling choice rather than a necessity of reinforcement learning.

Existing approaches primarily mitigate the symptoms of this formulation without revisiting its core assumption. Prior work compresses interaction histories~\citep{zhou2025mem1,luo2024efficient} or improves long-term credit assignment~\citep{liu2025agentic,zhai2025enhancing,xiong-etal-2024-watch}, but policies remain history-conditioned. Hierarchical reinforcement learning (HRL) introduces temporal abstraction and shows promise for LLM agents~\citep{hu2025divide}, yet current HRL methods still condition both high-level and low-level policies on accumulated interaction histories, inheriting the same long-context dependence they seek to alleviate.

To address the challenges discussed above, we propose \textbf{STEP-HRL} (Augmented \textbf{Step}-level \textbf{H}ierarchical \textbf{R}einforcement \textbf{L}earning), which rethinks long-horizon LLM agents from a \emph{progress-based} perspective. 
With completed high-level subtasks providing a \emph{global progress} of overall task, STEP-HRL introduces an additional \emph{local progress} module that accumulates subtask-relevant information at each timestep into a compact textual representation with controlled verbosity.
The low-level policy conditions exclusively on the current subtask, observation and the distilled local progress, enabling step-level decision making with constant-sized inputs.
Meanwhile, the local progress interacts with both the low-level and high-level policies, as well as with its own internal state, facilitating structured information transfer across hierarchical levels. We first perform behavior cloning on expert demonstrations to initialize policies, and then apply step-level offline RL for further optimization.

In summary, our contributions are as follows:
\begin{itemize}[leftmargin=10pt, itemsep=0pt, topsep=2pt]
  \item We propose STEP-HRL, a hierarchical framework that leverages a \textit{local progress} module to enable policies to condition on single-step transitions for LLM agents, eliminating the need to condition on full interaction histories.
  \item We propose a parameter-efficient two-stage training pipeline, where the high-level, low-level and local progress policies share a unified policy backbone, while being equipped with separate value networks for offline RL.
The model is first initialized via behavior cloning and subsequently fine-tuned with step-level offline optimization.
  \item Extensive experiments on the ScienceWorld and ALFWorld benchmarks demonstrate that our approach significantly improves both performance and generalization, validating the feasibility of step-level RL for LLM agents.
\end{itemize}

\section{Problem Formulation}
We formulate the agent operating in an interactive environment as a Partially Observable Markov Decision Process (POMDP), defined by the tuple $\langle \mathcal{C}, \mathcal{S}, \mathcal{A}, \mathcal{O}, \mathcal{T}, \mathcal{R} \rangle$.
Here, $\mathcal{C}$ denotes the instruction space specifying task goals, $\mathcal{S}$ is the latent environment state space, $\mathcal{A}$ is the action space, $\mathcal{O}$ is the observation space, $\mathcal{T} : \mathcal{S} \times \mathcal{A} \rightarrow \mathcal{S}$ represents the state transition dynamics, and $\mathcal{R} : \mathcal{S} \times \mathcal{A} \rightarrow \mathbb{R}$ is the reward function.
In the setting of LLM agents, $\mathcal{C}, \mathcal{A}, \mathcal{O}$ are expressed in natural language, while the environment state remains unobserved.
Given an instruction $c \in \mathcal{C}$, the agent interacts with the environment with policy $\pi_\theta$, the policy parameters $\theta$ are initialized from a pretrained LLM. The objective is to optimize the policy to maximize the expected discounted return $J(\pi) = \mathbb{E}_{\pi}\!\left[\sum_{t} \gamma^{t} r_t \right].$

\section{Method}
\begin{figure*}[t] \label{fig:overview}
  \includegraphics[width=0.99\linewidth]{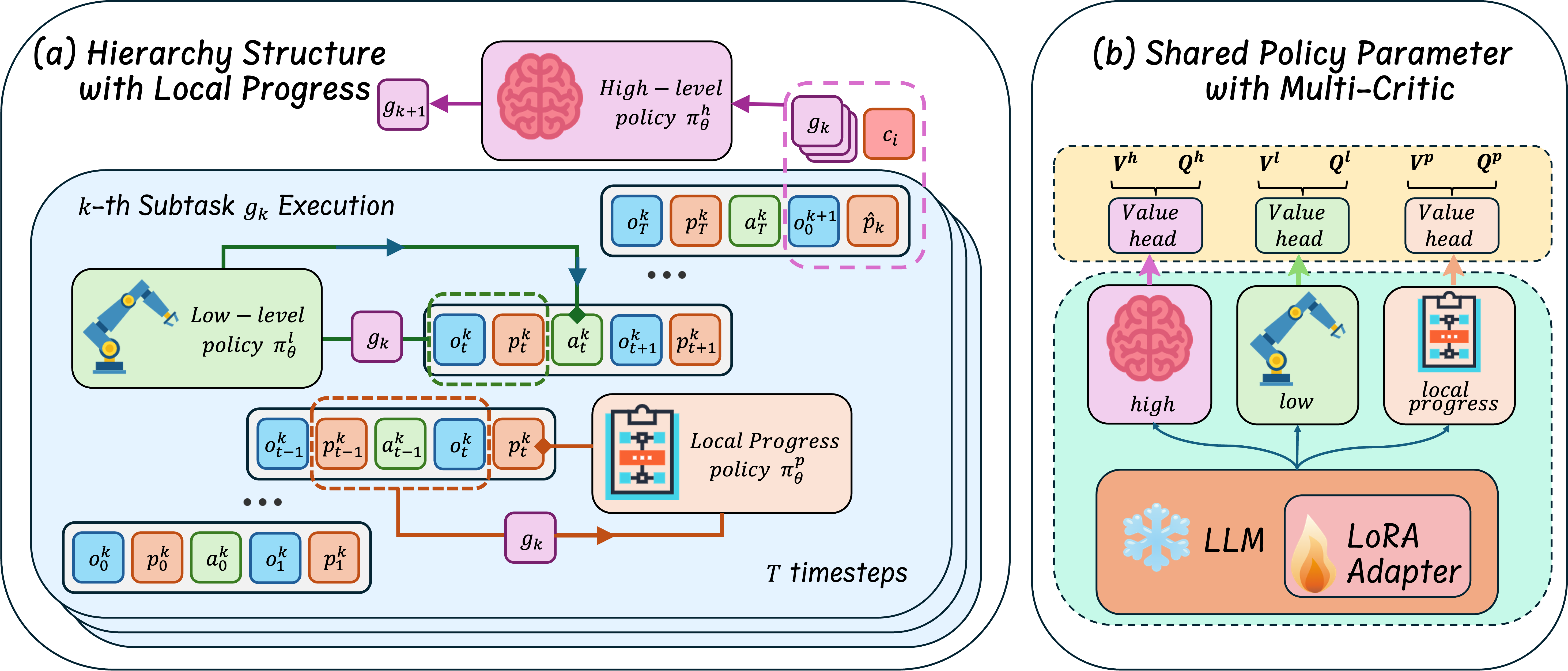}
  \caption {\textbf{(a)}: The pipeline of STEP-HRL. Local progress policy is responsible for producing a compact summary of local interaction history within each subtask. 
Specifically, the local progress policy $\pi^p_\theta$ depends on previous progress $p^k_{t-1}$, current subtask $g_k$, executed action $a^k_{t-1}$ and the resulting observation $o^k_t$ to the generate updated local progress $p^k_t$.
The low-level policy $\pi^l_\theta$ combines $p^k_t$ with observation $o^k_t$ and subtask $g_k$ to generate primitive actions. 
When current subtask $g_k$ terminates, its final local progress $\hat{p}_k$ is forwarded to the high-level policy $\pi^h_\theta$. Conditioned on the task instruction $c_i$, completed subtasks $G_k$, final local progress $\hat{p}_k$ and the initial observation $o^{k+1}_0$ of next subtask, $\pi^h_\theta$ generates the subsequent subtask.
\textbf{(b)}: The structure of our model. Three different policies share the same parameters, but equipped with different critic network respectively for offline RL training.
  }
\end{figure*}

\subsection{Step-Level Transitions with Local Progress Modeling}
Consider a commonly adopted history-conditioned RL formulation with hierarchical structures. Assume that all previous subtasks have been completed and a new subtask is to be generated.
The global interaction history up to time $t$ is denoted as
$\mathcal{H}_t = (c, o_0, g_0, a_0, \ldots, a_{t-1}, o_t)$, which concatenates the task instruction with past observations, subtasks, and actions. Conditioned on $\mathcal{H}_t$, the high-level policy generates the next subtask:
\begin{equation}
g_{k+1} \sim \pi^h_\theta(\cdot \mid \mathcal{H}_t).
\end{equation}

Given the $k$-th subtask $g_k$, the low-level policy operates at a finer temporal resolution. We denote the local interaction history as
$h^k_t = (o^k_0, a^k_0, \ldots, o^k_t)$,
which records observations and actions accumulated during subtask $g_k$.
Conditioned on the $g_k$ and $h^k_t$, the low-level policy produces primitive actions:
\begin{equation}
a^k_t \sim \pi^l_\theta(\cdot \mid g_k, h^k_t).
\end{equation}
The local history grows until the subtask terminates, after which the high-level policy is invoked again based on the updated global interaction history.

To enable step-level transitions, a key challenge is compactly representing both local and global interaction histories.
Intuitively, the sequence of completed subtasks $G_k = (g_0, g_1, \ldots, g_k)$ already serves as a concise summary of global task progress. Thus, the remaining problem is to compactly summarize the local interaction history within each subtask.
To this end, we introduce a \textit{local progress} policy $\pi_\theta^p$ to iteratively achieve this. 
At the beginning of the $g_k$, the local progress is initialized as $p^k_0 = \varnothing$, reflecting the absence of subtask-local interaction history. The local progress is then updated at each subsequent timestep according to:
\begin{equation}\label{lp_policy}
p^k_t \sim \pi_\theta^p(\cdot \mid g_k, a^k_{t-1}, o^k_t, p^k_{t-1}), \quad t > 0
\end{equation}
This design encourages $\pi_\theta^p$ to selectively extract subtask-relevant information from the previous progress $p^k_{t-1}$ and integrate it with the last executed action $a^k_{t-1}$ and its resulting observation $o^k_t$, yielding an updated local progress $p^k_t$.

With the $p^k_t$ capturing subtask-relevant local interaction information, the low-level policy can make decisions based on the augmented step-level transition $(o^k_t, p^k_t, a^k_t, \hat{r}^k_t, o^k_{t+1}, p^k_{t+1})$, where $\hat{r}^k_t$ is the intrinsic reward which equals $1$ if the current step successfully completes subtask $g_k$ and $0$ otherwise. This formulation enables step-level decision making without relying on the full interaction history within each subtask:
\begin{equation}\label{low_policy}
a^k_t \sim \pi^l_\theta(\cdot \mid g_k, p^k_t, o^k_t).
\end{equation}
It is worth noting that although $p^k_{t}$ already encodes information from current observation $o^k_t$, $o^k_t$ is typically most relevant for current action generation. To prevent $\pi^p_\theta$ from overlooking critical instantaneous information and to strengthen the sensitivity of $\pi^l_\theta$ to the current observation, we still explicitly include $o^k_t$ as an input to the low-level policy.

The local progress $p^k_t$ can also facilitate high-level subtask generation.
If we restrict high-level policy $\pi^h_\theta$ to condition solely on the completed subtasks $G_k$, it does not observe the detailed low-level progress. In this setting, the local progress $p^k_t$ bridges this information gap. For simplicity, we denote the final local progress at the termination of subtask $g_k$ as $\hat{p}_k$.
We pass the final progress $\hat{p}_{k-1}$ from the preceding subtask $g_{k-1}$ to the high-level policy. As a result, the step-level transition of high-level can be expressed as
$(\hat{p}_{k-1}, o^{k}_0, g_k, R_k, \hat{p}_k, o^{k+1}_0)$, where $R_k = \sum_{t} r^k_t$ is the accumulated extrinsic environment reward during subtask $g_k$, and the high-level policy generates the next subtask according to:
\begin{equation}\label{high_policy}
g_{k+1} \sim \pi^h_\theta(\cdot \mid c, G_{k}, \hat{p}_{k}, o^{k+1}_0).
\end{equation}

We adopt a parameter-efficient design across the three policies, $\pi^h_\theta$, $\pi^l_\theta$ and $\pi^p_\theta$ share the same parameters. This formulation facilitates efficient knowledge transfer across different decision levels, encourages consistent representations of task semantics and environment dynamics, and reduces the overall training and inference overhead. As a result, the three policies can be jointly optimized while maintaining clear functional specializatio.

\subsection{Behavior Cloning}
In interactive environments with specialized action and observation spaces, directly training LLM agents with RL often leads to poor sample efficiency. Moreover, since the three policies $\pi^h_\theta$, $\pi^l_\theta$ and $\pi^p_\theta$ must rapidly internalize their respective roles and output structures, we initialize the agent using expert demonstrations via behavior cloning.

We construct three expert demenstration datasets, $\mathcal{D}^h$, $\mathcal{D}^l$ and $\mathcal{D}^p$ based on the Eqs.~\eqref{lp_policy}, \eqref{low_policy} and \eqref{high_policy}. Specifically, We index tasks by $i \in [N]$, subtasks by $k \in [K_i]$, and within-subtask steps by $t \in [T_{i,k}]$. The datasets are organized as input-target pairs:
\begin{align}
\mathcal{D}^p &= \Big\{ \big((g_k, a^k_{t-1}, o^k_t, p^k_{t-1}),\; p^k_t\big) \Big\}, \label{d1}\\
\mathcal{D}^l &= \Big\{ \big((g_k, p^k_t, o^k_t),\; a^k_t\big) \Big\}, \label{d2}\\
\mathcal{D}^h &= \Big\{ \big((u_i, G_k, \hat{p}_k, o^{k+1}_0),\; g_{k+1}\big) \Big\}.\label{d3}
\end{align}

For notational convenience, we uniformly denote the policy input by $s$ and the action by $u$ across all three policies. Under this unified notation, behavior cloning optimizes each policy by:
\begin{equation}
\mathcal{L}_{\mathrm{BC}}(\theta)
= - \mathbb{E}_{(s,u)\sim \mathcal{D}} \left[ \log \pi_\theta (u \mid s) \right],
\end{equation}
where $\mathcal{D}$ denotes the corresponding expert demonstration dataset for each policy.
And the conditional log-likelihood $\log \pi_\theta(u \mid s)$ is computed autoregressively. Let $u = (w^{(1)}, \ldots, w^{(L)})$ denote the tokenization of the target output and $u^{(<\ell)} = (w^{(1)}, \ldots, w^{(\ell-1)})$ denote the preceding tokens. Then:
\begin{equation}\label{log_pi}
\log \pi_\theta(u \mid s)
= \sum_{\ell=1}^{L}  \log \pi_\theta\!\left(u^{(\ell)} \mid s, u^{(<\ell)}\right).
\end{equation}
This behavior cloning procedure serves as an effective initialization for subsequent RL stages. Empirically, even without further RL, our step-level behavior cloning alone achieves superior performance compared to existing baselines, as demonstrated in Section~\ref{experiments}.

\subsection{Step-Level Offline RL}
To further improve generalization, we collect an additional dataset $\tilde{\mathcal{D}}$ based on the behavior-cloned policies for offline optimization. We then combine the collected data with expert demonstrations to form the offline dataset
$\mathcal{D}_{\mathrm{r}} = \mathcal{D} \cup \tilde{\mathcal{D}}$, and optimize the policies on $\mathcal{D}_r$ using an actor-critic framework. 
We emphasize that the state $s$ corresponds to a \emph{single-step} state defined in Eqs.~\eqref{d1}–\eqref{d3}, instead of the full interaction history, which aligns with our step-level formulation.

\paragraph{Utterance-Level Implicit Value Learning.}
We implement the critic as an \emph{utterance-level} value estimator based on the hidden state of the last token. Concretely, given the final-token hidden state
\(
H \in \mathbb{R}^{B \times d},
\)
the critic attaches two lightweight MLP heads that output scalar predictions for the state-value function \(V_\psi(s)\) and the action-value function \(Q_\phi(s,u)\), respectively.

Following the implicit value learning paradigm introduced in Implicit Q-Learning (IQL) and its language adaptation ILQL~\cite{kostrikov2021IQL,snellILQL}, we jointly learn the \(Q_\phi\) and the \(V_\psi\) using  step-level transitions $(s,u,r,s')$ rather than full trajectories. The Q-function is trained by minimizing a TD regression loss bootstrapped from the value function:
\begin{equation}
\begin{aligned}
\mathcal{L}_{Q}(\phi)=
\mathbb{E}_{(s,u,r,s') \sim \mathcal{D}_{\mathrm{r}}}
\Big[
(r
&+ \gamma V_{\bar{\psi}}(s') \\
&- Q_{\phi}(s,u))^2
\Big],
\end{aligned}
\label{eq:critic_td}
\end{equation}
where \(V_{\bar{\psi}}\) denotes a softly updated target value network used to stabilize training~\cite{haarnoja2018soft}.
To approximate the constrained Bellman optimality operator without explicitly maximizing over actions, the value function \(V_\psi(s)\) is trained using \emph{expectile regression}. Specifically, \(V_\psi(s)\) is optimized to regress toward the action-value estimates under an asymmetric squared loss:
\begin{equation}
\mathcal{L}_{V}(\psi)
=
\mathbb{E}_{(s,u) \sim \mathcal{D}_{\mathrm{r}}}
\left[
L_2^{\tau}
\big(
Q_{\bar{\phi}}(s,u) - V_\psi(s)
\big)
\right].
\end{equation}
where we define \(d = Q_{\bar{\phi}}(s,u) - V_\psi(s)\) and $L_2^{\tau}(d) = \left| \tau - \mathbf{1}(d < 0) \right| d^2$
is the expectile loss with expectile parameter \(\tau \in (0,1)\) By choosing \(\tau > 0.5\), the value function approximates an upper expectile of the maximum.
This mechanism implicitly biases learning toward high-value actions, enabling stable offline RL without an explicit policy optimization step.

\paragraph{Implicit Policy Improvement via Advantage-Weighted Regression.}
The policy is trained to assign higher likelihood to actions with higher estimated value under the learned critic. Concretely, given step-level transitions from the offline dataset $\mathcal{D}_r$, the policy is optimized by regressing toward actions favored by the critic. We employ an advantage-weighted regression objective: 
\begin{equation}
\begin{aligned}
\mathcal{L}_{A}(\theta)
= - \mathbb{E}\Big[
\exp\!\Big(
\frac{A(s, u)}{\beta}
\Big)
\, \log \pi_\theta(u \mid s)
\Big], \\
\text{where}\quad
A(s, u) = Q_\phi(s, u) - V_\psi(s).
\end{aligned}
\end{equation}
The computation of $\log \pi_\theta(u \mid s)$ follows the same autoregressive procedure as in Eq.~\eqref{log_pi}.
Following prior offline RL methods~\cite{peng2019AWR,kostrikov2021IQL,nair2020awac}, we employ an exponential advantage-weighted objective.
The temperature parameter $\beta$ controls the sharpness of the weighting and balances policy improvement strength and training stability in our setting.

We apply the offline RL procedure to all three policies, $\pi^h_\theta$, $\pi^l_\theta$, and $\pi^p_\theta$, each equipped with a separate critic network while sharing the same policy parameters. This design provides task-specific value supervision at different levels of abstraction, while shared policy parameters facilitate effective knowledge transfer across hierarchical decisions.
Consequently, the model captures complementary decision patterns across levels and achieves more consistent and sample-efficient learning across tasks.

\section{Experiments}\label{experiments}

\subsection{Experimental Settings}

\paragraph{Benchmarks and Datasets.}
We evaluate our approach on two challenging benchmarks:
\begin{itemize}[leftmargin=10pt, itemsep=0pt, topsep=2pt]
    \item \textbf{ScienceWorld}~\citep{wang2022scienceworld} is a text-based interactive benchmark with 30 science task families (e.g., physics, chemistry, biology), each containing many parameterized variants, yielding hundreds to thousands of tasks that require multi-step reasoning and experimentation.
    \item \textbf{ALFWorld}~\citep{shridhar2020alfworld} is a household task benchmark aligned with ALFRED, comprising134 language-conditioned tasks across 6 task types (e.g., pick-and-place, cleaning, heating), focusing on long-horizon action planning.
\end{itemize}

For dataset construction, we generate hierarchical annotations, including subtasks and local progress signals from expert trajectories using \textsc{DeepSeek}. The prompts used for subtask decomposition and progress generation are provided in the Appendix~\ref{Benchmarks and Datasets}. 
For offline RL, we collect additional trajectories using policies initialized via behavior cloning. Following the experimental setup of GLIDER~\cite{hu2025divide}, we adopt a trajectory mixture ratio of $1\!:\!2$, which was identified as the most effective setting in their study.

\paragraph{Models and Baselines.}
We evaluate STEP-HRL on three outstanding open source models: \textbf{Mistral-7B}~\cite{Jiang2023Mistral7}, \textbf{Gemma-7B}~\cite{team2024gemma} and \textbf{Llama3-8B}~\cite{meta2024llama3}.

We compare against the following baselines:
1) \textbf{ReAct}~\citep{yao2022react}, a prompting framework that interleaves reasoning traces and environment actions in a Thought--Action--Observation loop.
2) \textbf{Reflexion}~\citep{shinn2023reflexion}, which improves subsequent trials by storing self-reflective feedback in an episodic memory.
3) \textbf{SwiftSage}~\citep{lin2023swiftsage}, a dual-process agent that combines a behavior-cloned action model with an LLM-based planner for interactive tasks.
4) \textbf{ETO}~\citep{song-etal-2024-eto}, which iteratively collects contrastive (failure/success) trajectories and optimizes the policy via DPO~\cite{rafailov2023dpo}.
5) \textbf{WKM}~\citep{qiao2024WKM}, which augments planning with a parametric world knowledge model that provides task priors and dynamic state knowledge.
6) \textbf{GLIDER}~\citep{hu2025divide}, an offline HRL framework that decomposes complex tasks and learns high-level and low-level policies for decision making. We also report results of \textbf{ChatGPT} (\texttt{gpt-3.5-turbo-0125}) and \textbf{GPT-4} (\texttt{gpt-4-32k-0613}) for comparison by referencing previously published results~\citep{qiao2024WKM}.

\paragraph{Training Details.}
All fine-tuning baselines and our method are fine-tuned using LoRA~\citep{hu2022lora}.
For behavior cloning, we train the policies for 5 epochs with a learning rate of $1\times10^{-4}$ and a batch size of 128.
During the offline RL stage, we train for 3 epochs, using learning rates of $1\times10^{-5}$ and $1\times10^{-4}$ for the actor and critic, respectively, with a batch size of 256.
All models are optimized using AdamW~\citep{loshchilov2017decoupled} optimizer.
All experiments are conducted on 8 NVIDIA A100 80G GPUs.
Detailed hyperparameters and additional experimental settings are provided in Appendix~~\ref{Training Details}.

\subsection{Results}
\input{table/main_result.tex}\label{tab:main_result}
\paragraph{Main results.} 
Table~\ref{tab:main_result} reports the evaluation results of \textsc{STEP-HRL} across three backbone models on the ScienceWorld and ALFWorld benchmarks.
Across all settings, \textsc{STEP-HRL} consistently outperforms strong prior baselines on both seen and unseen tasks.
On ALFWorld, \textsc{STEP-HRL} achieves near-saturated performance, with success rates exceeding 90\% across different backbone models.
On ScienceWorld, \textsc{STEP-HRL} also yields consistent and substantial improvements over existing methods, demonstrating its effectiveness in more challenging and diverse environments.
Notably, the performance gap between backbones is substantially reduced, indicating strong robustness and scalability of our proposed framework.

\paragraph{Performance across Different Model Scales.}
\input{table/model_scale.tex}\label{tab:model_scale}
Table~\ref{tab:model_scale} summarizes the performance of \textsc{STEP-HRL} across different model scales. Overall, performance improves steadily as model capacity increases, with larger backbones achieving better performance. Notably, even smaller models such as Llama-1B and Llama-3B demonstrate competitive performance, particularly on ALFWorld. This trend suggests that \textsc{STEP-HRL} is exceptionally effective even across a wide range of model scales, while additional model capacity further enhances robustness and generalization, especially on more challenging ScienceWorld tasks.

\subsection{Ablation Studies}
\begin{figure} 
  \includegraphics[width=\linewidth]{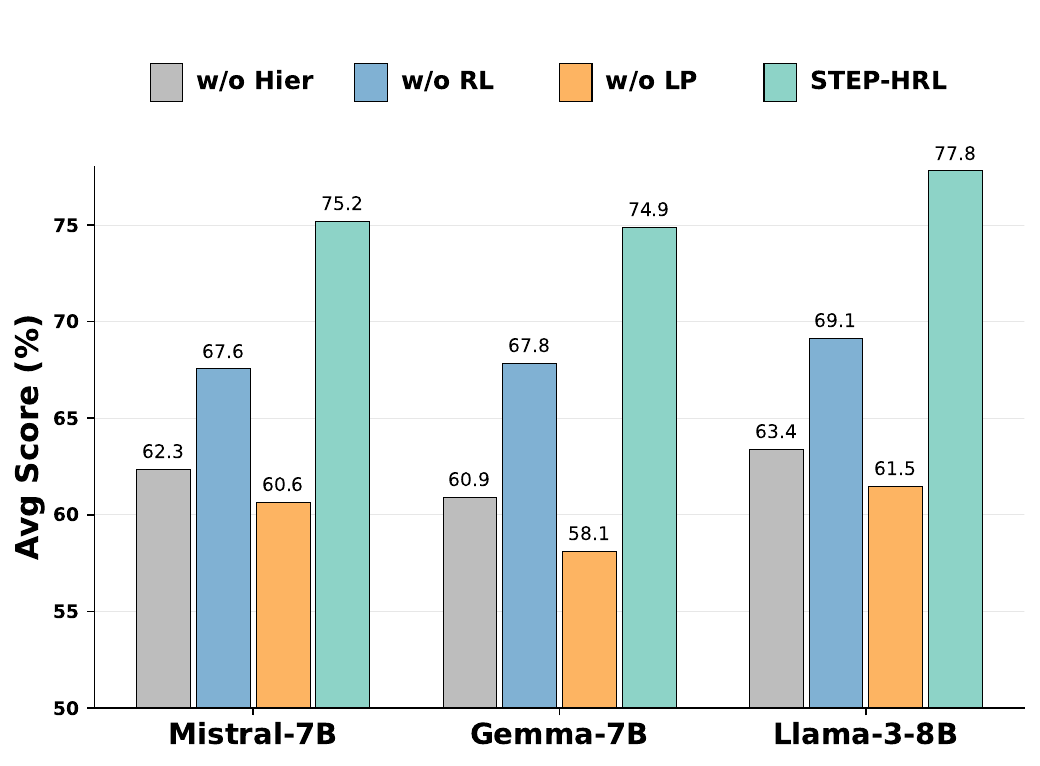}
  \caption {Ablation study of \textsc{STEP-HRL} on unseen ScienceWorld tasks with different backbone models. \texttt{w/o LP} denotes removing the local progress policy, \texttt{w/o Hier} denotes removing the hierarchical structure, and \texttt{w/o RL} denotes removing the offline RL stage,  reducing the training procedure to behavior cloning only.
  \label{fig:ablation}
}\end{figure}

As shown in Figure~\ref{fig:ablation}, we examine the effectiveness of key components in STEP-HRL. 
We consider three variants: 1) \texttt{w/o LP}, which removes the local progress policy, forcing the low-level and high-level policies to condition on $(g_k, o^k_t)$ and $(c_i, G_{k}, o^{k+1}_0)$ respectively; 
2) \texttt{w/o Hier}, which eliminates the hierarchical structure and directly relies on the local progress module to summarize the global interaction history;
3) \texttt{w/o RL}, which omits the offline RL stage and reduces training to behavior cloning only.
All variants are trained using the same step-level data for a fair comparison.

Across all settings, alternative variants consistently lead to degraded performance.
Most notably, the local progress module plays a central role by condensing subtask-relevant interaction information into a compact summary Without this module, many states become indistinguishable, making credit assignment and policy optimization significantly more challenging.
The hierarchical structure further contributes by decomposing complex tasks into manageable subtasks, which alleviates the burden on the local progress module and prevents it from being overwhelmed when summarizing long-horizon interactions.
Finally, the offline RL stage refines the policies beyond behavior cloning, improving generalization to unseen tasks through value-guided policy updates.
Collectively, these results highlight the complementary roles of all components and underscore the importance of the proposed design in STEP-HRL.

\subsection{Analysis on Efficiency}
\begin{figure} 
  \includegraphics[width=\linewidth]{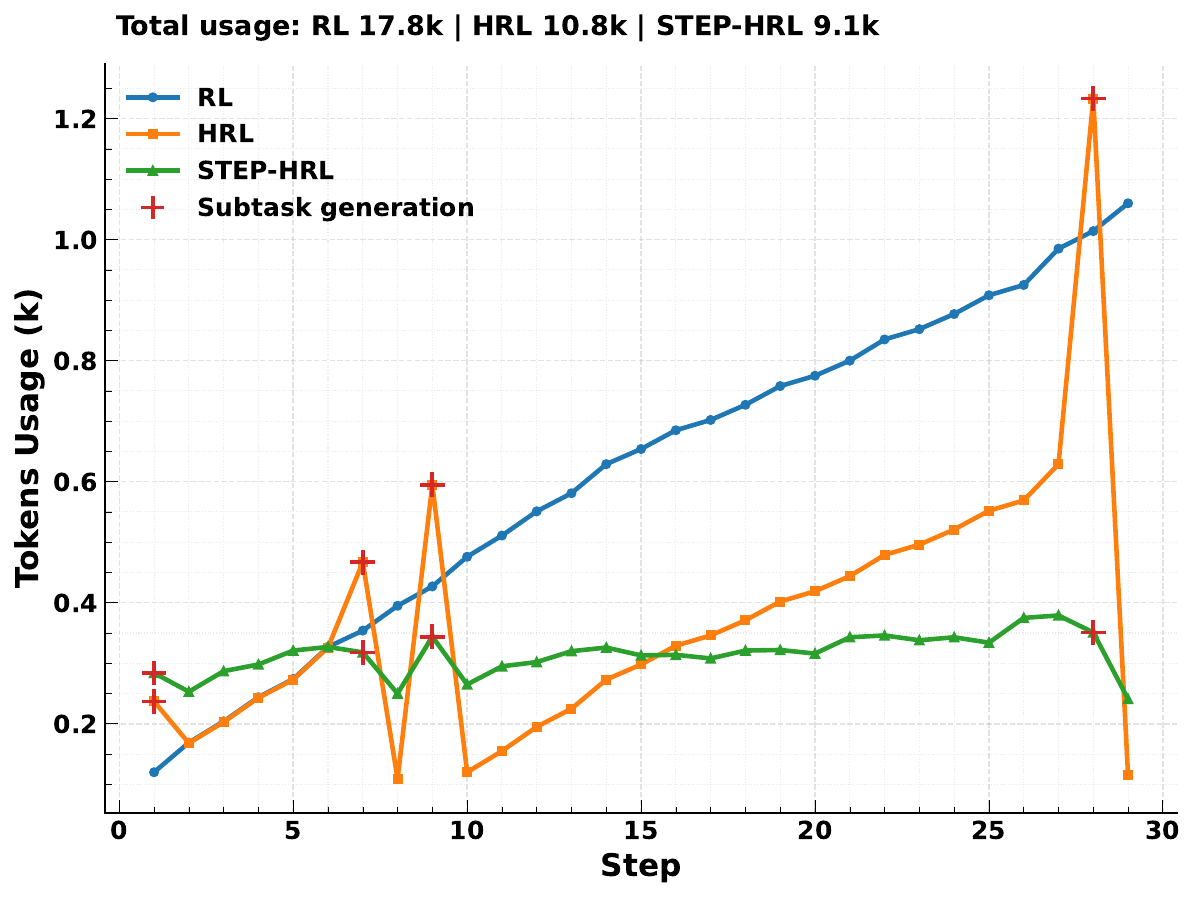}
  \caption {Simulated per-step token usage on ALFWorld \texttt{pick\_two\_obj\_and\_place} task
under identical observation and action sequence across three RL paradigms. 
\label{fig:token}
}\end{figure}

\paragraph{Token Usage Efficiency.}
As shown in Figure~\ref{fig:token}, we analyze the token efficiency of STEP-HRL in comparison with standard RL (i.e., a normal ReAct-style agent) and HRL on an ALFWorld task.
For a fair comparison, all methods are evaluated under the same observation and action sequence. The task is decomposed into four subtasks and requires a total of 29 environment steps to complete.

Standard RL incurs steadily increasing per-step token costs as the interaction progresses, since the policy repeatedly conditions on an ever-growing interaction history.
Although HRL reduces the token usage by decomposing the task into subtasks, it exhibits substantial variability, with pronounced spikes at subtask generation steps where long accumulated contexts are processed.
Such high variance in input sequence not only increases inference latency, but also leads to inefficient training.
In particular, GPU memory allocation must accommodate peak input lengths induced by high-level samples, resulting in underutilization during most steps and reduced overall training efficiency.

In contrast, STEP-HRL maintains an approximately constant per-step token usage.
By leveraging compact summaries of both global task progress and local subtask progress, STEP-HRL avoids conditioning on full interaction histories.
This design effectively bounds the per-step inference cost, yielding the lowest average token usage with minimal variance.
Overall, the result highlights the advantage of STEP-HRL in enabling predictable, efficient and well-balanced inference and training behavior, which is particularly important for long-horizon interactive environments.

\paragraph{Wall-Clock Inference Efficiency.}
We further conduct analysis on wall-clock inference latency. As shown in Table~\ref{tab:latency-analysis}, STEP-HRL is slightly slower than normal agent framework in early steps. This is because STEP-HRL contains three policies and therefore performs more forward passes per step. Under KV-cache decoding, this introduces additional prefills, and local-progress generation typically produces more tokens than primitive actions.

However, as the interaction history grows, the latency of the history-conditioned baseline increases, whereas STEP-HRL remains stable. This suggests that STEP-HRL is less sensitive to history length and scales more favorably in long-horizon settings.

\input{table/latency.tex}

\subsection{Analysis on Offline RL}
\begin{figure} 
  \includegraphics[width=\linewidth]{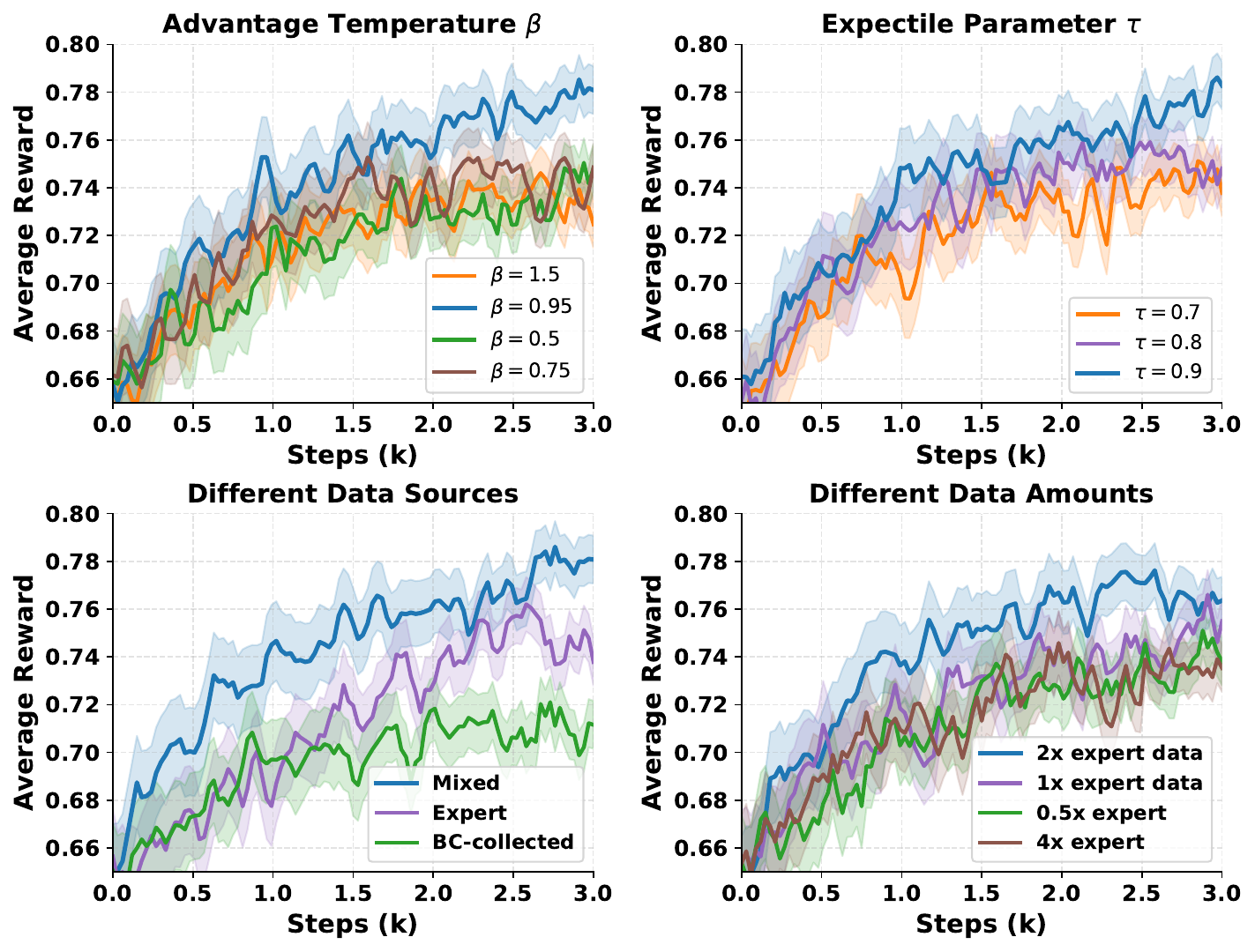}
  \caption {Offline RL sensitivity analysis with respect to advantage temperature $\beta$, expectile parameter $\tau$, data sources, and data amounts across training.
}
\label{fig:rl}
\end{figure}

Figure~\ref{fig:rl} presents a sensitivity analysis of the proposed offline RL procedure with respect to key factors.
We observe that the advantage temperature $\beta$ plays a critical role in balancing update aggressiveness and stability: among the tested values, $\beta=0.95$ consistently yields the highest final performance, while both smaller and larger temperatures lead to inferior convergence.
Similarly, varying the expectile parameter $\tau$ reveals that a higher expectile (e.g., $\tau=0.9$) provides more effective value estimation and results in stronger policy improvement compared to lower settings.

We further analyze the impact of data sources used for offline RL.
Training on mixed datasets that combine expert demonstrations with BC-collected trajectories consistently outperforms using either expert-only or BC-collected data alone.
While BC-collected data contains informative failure trajectories that are valuable for policy improvement, it also introduces lower-quality and noisier samples compared to expert demonstrations.
We also study the effect of scaling the amount of training data.
Increasing data size improves performance up to a point, with twice the expert data achieving the best overall results.
Beyond this regime, additional data yields diminishing returns, as excessively large datasets introduce redundant or low-quality samples that hinder stable learning.

\section{Related Work}
\paragraph{LLM Agents.}
With their strong semantic understanding and emergent reasoning abilities, large language models (LLMs) have been explored as autonomous agents for decision making in complex and interactive environments~\cite{guo2024large,wang2024survey}. 
Early studies primarily adopt prompt-based formulations, where agents generate intermediate reasoning traces to support multi-step decisions, such such as Chain-of-Thought~\cite{wei2022chain}, ReAct~\cite{yao2022react}, Reflexion~\cite{shinn2023reflexion} and their variants~\cite{yao2023tree,wang2022self}. 
Subsequent work augments LLM agents with additional system components, including tool use~\cite{schick2023toolformer,qin2023toolllm,wu2024avatar,li2025meco}, memory mechanisms~\cite{zhang2025survey,sarch2024vlm,xu2025mem}, and multi-agent coordination~\cite{chen2024agentverse,bo2024reflective,estornell2024acc}.
Beyond architectural augmentation, another line of work focuses on grounding LLM agents through learning from expert demonstration via fine tuning (behavior cloning), demonstrating strong gains ~\cite{zeng2024agenttuning,chen2023fireact,yin2023lumos, chen2024agent}.
However, these approaches heavily rely on high-quality expert data and trajectory-level supervision, and their performance degrades in long-horizon decision making and complex interactive tasks due to limited exploration and severe distribution shift..

\paragraph{Reinforcement Learning in LLM Agents.} Reinforcement learning (RL) has achieved notable success in aligning and improving large language models~\cite{ouyang2022training, shao2024deepseekmath}, and has also proven effective for training LLM agents through explicit reward and penalty mechanisms.
Most prior work adopts an interaction-driven pipeline, where an LLM agent receives goal-directed feedback from the environment and is fine-tuned using RL algorithms such as PPO~\cite{schulman2017proximal,zhai2024fine,szot2023large,peiyuan2024agile}.
Preference-based methods, such as ETO~\cite{song-etal-2024-eto}, further collect contrastive trajectory pairs from environment interactions and update LLM policies via preference optimization objectives like DPO~\cite{rafailov2023dpo}.
To achieve fine-grained reinforcement learning signals instead of optimizing the full trajectories, \citet{wen2024reinforcing} decompose RL objectives to provide action-level feedback for LLM agents, while GiGPO~\cite{feng2025group} hierarchically estimates step-level advantages to improve training efficiency.
Despite their effectiveness, most existing RL-based LLM agents depend on full interaction histories for decision making, where the increasing context length poses significant challenges for credit assignment and computational efficiency.

To mitigate this limitation, several works have explored HRL frameworks such as EPO~\cite{zhao-etal-2024-epo} and GLIDER~\cite{hu2025divide} decompose complex tasks into subtasks and learn coordinated high-level and low-level policies.
However, even docompsing or fine-grained optimization, these methods still rely on history-conditioned policies, resulting in inefficient credit assignment and high computational overhead.

\section{Conclusion}
In this paper, we proposed STEP-HRL, a innovative framework that enables efficient step-level learning for LLM agents without relying on full interaction histories.
STEP-HRL decomposes tasks into a hierarchical structure and introduces a local progress module to summarize subtask-relevant information, allowing both high-level and low-level policies to operate on compact, step-level state representations.
Empirical results on ScienceWorld and ALFWorld demonstrate that STEP-HRL consistently improves performance and generalization.
Overall, STEP-HRL provides a practical and scalable approach for training LLM agents. We believe that step-level abstraction with structured progress summaries offers a promising direction for improving both efficiency and robustness in future LLM agent research.

\section*{Limitations}
Despite the effectiveness of STEP-HRL, it still has several limitations worth noting:
\begin{itemize}[leftmargin=10pt, itemsep=0pt, topsep=2pt]
  \item STEP-HRL highly relies on high-quality expert demonstrations.
In particular, the construction of step-level data requires carefully designed subtask and local progress.
Designing and curating them can be non-trivial in practice, especially for complex environments with ambiguous subtask boundaries or poorly defined progress signals. 
This reliance may limit the applicability of STEP-HRL in domains where expert data or structured supervision is scarce.
  \item In our implementation, subtask termination is predicted jointly with primitive actions, such that each low-level output includes both an action and a termination indicator.
This design may result in inaccurate termination decisions, including premature termination or delayed subtask completion.
Such errors can degrade the quality of collected transitions and introduce bias into critic value estimation, which in turn may cause misalignment between high-level planning and low-level execution during inference.
\end{itemize}

\section*{Acknowledgments}
The work was supported by the \href{https://data.crossref.org/fundingdata/funder/10.13039/501100001809}{National Natural Science Foundation of China} (Grant No. U22B2019), \href{https://data.crossref.org/fundingdata/funder/10.13039/501100004543}{China Scholarship Council} (No. 202406470057) and Beijing University of Posts and Telecommunications-China Mobile Communications Group Co., Ltd. Joint Institute.

\bibliography{custom}

\clearpage
\appendix

\section{Benchmarks and Datasets} \label{Benchmarks and Datasets}
\subsection{Benchmarks}
We evaluate our approach on two widely used language-based interactive decision-making benchmarks: ScienceWorld and ALFWorld.
\begin{itemize}[leftmargin=10pt, itemsep=0pt, topsep=2pt]
  \item \textbf{ScienceWorld}~\citep{wang2022scienceworld} is a text-based environment designed for science experimentation.
It consists of 30 tasks spanning 10 categories, where agents are required to demonstrate scientific reasoning through interactive exploration.
The environment provides dense rewards at each step, with values ranging from 0 to 1, reflecting incremental task progress.
  \item \textbf{ALFWorld}~\citep{shridhar2020alfworld} simulates household environments that involve navigation and object manipulation.
In contrast to ScienceWorld, ALFWorld adopts a sparse reward setting, where an agent receives a reward of 1 only upon successful task completion and 0 otherwise.
\end{itemize}

Both ScienceWorld and ALFWorld are evaluated under two settings: \emph{Seen} and \emph{Unseen}.
The Seen split contains in-distribution tasks that follow the similar task and variations as those observed during training, and is used to evaluate in-distribution performance.
In contrast, the Unseen split consists of out-of-distribution task variations with novel mechanism or object, and is used to assess the generalization ability. Dataset statistics for all splits are summarized in Table~\ref{tab:dataset_statistics}.
\input{table/task_statistics.tex}

\subsection{Datasets}
\paragraph{Expert Dataset.} 
For subtask and local-progress generation, we employ a combination of rule-based heuristics and the \textsc{DeepSeek} model.
For ScienceWorld, due to the substantial structural diversity across tasks, we adopt task-specific prompts and subtask decomposition strategies tailored to each task category.
In contrast, for ALFWorld, we design a unified prompt that guides \textsc{DeepSeek} to generate both subtask and local-progress fields in a consistent manner.
Figure~\ref{fig:lp_prompt} presents the prompt used for generating local progress.

After generating subtask and local progress fields, we construct the SFT (BC) datasets for all three policies. The data structure of these datasets are shown below, and the prompts used during training and inference are provided in Figure~\ref{fig:prompt}.

\input{box/traj.tex}


\paragraph{Offline RL Dataset.}
We construct the RL dataset by combining expert demonstrations with trajectories collected from behavior-cloned policies.

During data collection, we adopt different sampling temperatures for different policy components to balance exploration and action validity.
Specifically, for the high-level policy and the local-progress policy, we set the sampling temperature to \textbf{1.0} to encourage diverse subtask sequences and reasoning paths.
In contrast, the low-level policy generates primitive actions that must strictly conform to the environment’s action format and input constraints.
To avoid producing invalid or malformed actions, we therefore set the sampling temperature of the low-level policy to \textbf{0}, ensuring deterministic and well-formed action generation.

The final offline RL dataset consists of both expert data and policy-collected trajectories, with an approximate ratio of \textbf{1:2}.
Among the collected trajectories, around \textbf{25\%} correspond to unsuccessful episodes.
Including such unsuccessful data enables the model to observe negative outcomes and learn to penalize suboptimal actions, which is beneficial for stable offline policy learning.

\clearpage
\onecolumn
\input{box/lp_prompt.tex}
{\captionsetup{type=figure,skip=2pt}
\captionof{figure}{Full prompt specification for local progress annotation on the ALFWorld expert dataset (\textsc{DeepSeek}).
\label{fig:lp_prompt}}}

\input{box/prompt.tex}
{\captionsetup{type=figure,skip=2pt}
\captionof{figure}{The prompt used in training and inference stages.
\label{fig:prompt}}}

\clearpage
\twocolumn
\section{Training Details}
\label{Training Details}
\paragraph{Models.}
We conduct our main experiments using the following instruction-tuned large language models:
\begin{itemize}\setlength{\itemsep}{0.2em}
  \item \texttt{mistralai/Mistral-7B-Instruct-v0.2}
  \item \texttt{google/gemma-1.1-7b-it}
  \item \texttt{meta-llama/Meta-Llama-3-8B-Instruct}
\end{itemize}

For scalability experiments across different model sizes, we additionally evaluate:
\begin{itemize}\setlength{\itemsep}{0.2em}
  \item \texttt{meta-llama/Llama-3.2-1B-Instruct}
  \item \texttt{meta-llama/Llama-3.2-3B-Instruct}
\end{itemize}

\paragraph{Hyperparameters.}
The details of all hyperparameters are summarized in Table~\ref{tab:para}.
We adopt LoRA for parameter-efficient fine-tuning across all models. During the behavior cloning (BC) stage, models are trained for 5 epochs using the AdamW optimizer with a learning rate of $1\times10^{-4}$. Training is performed with a total batch size of 128, achieved via a per-device batch size of 8 and 2 gradient accumulation steps, which balances computational efficiency and training stability.

For the offline reinforcement learning stage, we train the model for 3 epochs, using separate learning rates for the actor ($1\times10^{-5}$) and critic ($1\times10^{-4}$) . The target critic is updated via soft updates with coefficient $\tau_1=0.2$, and the discount factor is set to $\gamma=0.99$. To further stabilize training, we first warm up the critic for 100 steps before jointly optimizing the actor and critic. We employ an advantage-weighted objective with weighting factor $\beta=0.95$, and adopt expectile regression with parameter $\tau_2=0.95$ for value learning.

During trajectory sampling, we use a sampling temperature of 0.7 for the high-level and local-progress policies to encourage diverse reasoning paths, while setting the temperature to 0 for the low-level policy to ensure deterministic and valid primitive action generation. We further constrain text generation by allowing a maximum of 32 tokens for the high-level and low-level policies, and 150 tokens for the local-progress policy.

For evaluation, we impose a maximum of 50 environment steps per episode for both ALFWorld and ScienceWorld tasks, ensuring a consistent evaluation budget across benchmarks.

\input{table/hyperparameter_one.tex}

\clearpage
\onecolumn
\section{Evaluation Details}
We further report model performance on each individual task family. Since the result distributions are similar across different backbone models, we only present results for Llama-3-8B in Tables~\ref{tab:eval_alfworld} and \ref{tab:eval_scienceworld}.

\input{table/eval_alfworld.tex}
\input{table/eval_scienceworld.tex}

\clearpage
\section{Case Study}
\vspace{-1.5em}
\input{box/case1.tex}
\input{box/case2.tex}
{\captionsetup{type=figure,skip=2pt}
\captionof{figure}{The trajectory of STEP-HRL on ScienceWorld \texttt{boil} task.
\label{fig:lp_prompt}}}

\end{document}

%% file: table/main_result.tex
\begin{table*}[!t]
\setlength{\tabcolsep}{5pt}   
\renewcommand\arraystretch{0.8}
\caption{\small
\textbf{Main Results.}
We report performance comparisons for three backbone models on the ScienceWorld and ALFWorld benchmarks. 
{\small \faToggleOff} corresponds to prompt-based methods without updating model parameters, and {\small \faToggleOn} corresponds to LoRA-based fine-tuning approaches. 
\textcolor{red}{$\uparrow$} indicates the relative improvement of STEP-HRL compared to the best results among the baselines.
}
\centering
\scalebox{0.96}{
\begin{tabular}{c|l|cc|cc}
\toprule

{\multirow{2}{*}{\textbf{Backbone}}} 
&{\multirow{2}{*}{\textbf{Method}}} 
& \multicolumn{2}{c|}{\textbf{ScienceWorld}} 
& \multicolumn{2}{c}{\textbf{ALFWorld}} \\
\cmidrule{3-4}
\cmidrule{5-6}
& & Seen & Unseen & Seen & Unseen \\
\midrule
GPT-3.5-Turbo &
\multirow{2}{*}{{\small \faToggleOff} \textsc{ReAct}} &
8.57 & 5.97 & 15.41 & 13.99 \\
GPT-4 &
&
44.29 & 38.05 & 67.32 & 65.09 \\
\midrule
\multirow{6}{*}{\makecell{Mistral-7B}} &
{\small \faToggleOff}   ReAct       & 20.72 & 17.65 & 7.86  & 5.22  \\
& {\small \faToggleOff} Reflexion   & 21.07 & 18.11 & 11.56 & 6.00  \\
& {\small \faToggleOff} SwitchSage  & 48.40 & 45.25 & 30.29 & 26.52 \\
& {\small \faToggleOn}  ETO         & 58.17 & 51.85 & 66.84 & 71.43 \\
& {\small \faToggleOn}  WKM         & 62.12 & 53.62 & 73.57 & 76.87 \\
& {\small \faToggleOn}  GLIDER      & 67.31 & 65.14 & 70.02 & 74.83 \\
\cmidrule{2-6}
& {\small \faToggleOn} \textbf{\ours} 
& \textbf{80.28} {\small (\textcolor{red}{$\uparrow 19.27\%$})} 
& \textbf{75.21} {\small (\textcolor{red}{$\uparrow 15.46\%$})} 
& \textbf{96.43} {\small (\textcolor{red}{$\uparrow 31.07\%$})} 
& \textbf{97.01} {\small (\textcolor{red}{$\uparrow 26.20\%$})} \\

\midrule
\multirow{6}{*}{\makecell{Gemma-7B}} &
{\small \faToggleOff}   ReAct       & 3.58  & 3.51  & 6.43  & 2.24  \\
& {\small \faToggleOff} Reflexion   & 4.94  & 3.93  & 7.14  & 2.99  \\
& {\small \faToggleOff} SwitchSage  & 33.43 & 30.90 & 8.23  & 5.72  \\
& {\small \faToggleOn}  ETO         & 50.44 & 47.84 & 66.43 & 68.66 \\
& {\small \faToggleOn}  WKM         & 53.68 & 49.24 & 70.71 & 70.40 \\
& {\small \faToggleOn}  GLIDER      & 63.67 & 58.50 & 72.12 & 70.88 \\
\cmidrule{2-6}
& {\small \faToggleOn} \textbf{\ours} 
& \textbf{78.89} {\small (\textcolor{red}{$\uparrow 24.02\%$})} 
& \textbf{74.08} {\small (\textcolor{red}{$\uparrow 26.63\%$})} 
& \textbf{97.86} {\small (\textcolor{red}{$\uparrow 35.69\%$})} 
& \textbf{97.76} {\small (\textcolor{red}{$\uparrow 37.92\%$})} \\

\midrule
\multirow{6}{*}{\makecell{Llama-3-8B}} &
{\small \faToggleOff}   ReAct       & 24.76 & 22.66 & 2.86  & 3.73  \\
& {\small \faToggleOff} Reflexion   & 27.23 & 25.41 & 4.29  & 4.48  \\
& {\small \faToggleOff} SwitchSage  & 42.22 & 40.58 & 20.39 & 10.78 \\
& {\small \faToggleOn}  ETO         & 57.90 & 52.33 & 64.29 & 64.18 \\
& {\small \faToggleOn}  WKM         & 60.12 & 54.75 & 68.57 & 65.93 \\
& {\small \faToggleOn}  GLIDER      & 77.43 & 68.34 & 71.56 & 75.38 \\
\cmidrule{2-6}
& {\small \faToggleOn} \textbf{\ours} 
& \textbf{81.57} {\small (\textcolor{red}{$\uparrow 5.35\%$})} 
& \textbf{77.81} {\small (\textcolor{red}{$\uparrow 13.86\%$})} 
& \textbf{97.14} {\small (\textcolor{red}{$\uparrow 35.75\%$})} 
& \textbf{97.76} {\small (\textcolor{red}{$\uparrow 29.69\%$})} \\

\bottomrule
\end{tabular}
}
\label{tab:main_results}
\end{table*}

%% file: table/model_scale.tex
\begin{table}[htb]
\setlength{\tabcolsep}{6pt}
\renewcommand\arraystretch{1}
\caption{Performance of STEP-HRL on ScienceWorld and ALFWorld across model scales.}
\centering
\scalebox{0.9}{
\begin{tabular}{l|cc|cc}
\toprule
\multirow{2}{*}{\textbf{Model}} 
& \multicolumn{2}{c|}{\textbf{ScienceWorld}} 
& \multicolumn{2}{c}{\textbf{ALFWorld}} \\
\cmidrule(l){2-3} \cmidrule(l){4-5}
& Seen & Unseen & Seen & Unseen \\
\midrule
Llama-1B & 51.88 & 49.78  & 89.86 & 89.60 \\
Llama-3B & 65.31 & 61.79 & 94.29 & 94.00 \\
Llama-8B & 80.57 & 77.81 & 96.43 & 97.76 \\
\bottomrule
\end{tabular}}
\end{table}


%% file: table/latency.tex
\begin{table}[!ht]
\caption{Per-step inference latency on a single A100 GPU in the ALFWorld Pick\&Place. 
$\star$ denotes subtask generation, which introduces an additional forward pass.}
\label{tab:latency-analysis}
\centering
\begin{tabular}{lcc}
\hline
\textbf{Step} & \textbf{Ours (s)} & \textbf{Normal (s)} \\
\hline
1  & 1.8 $\pm$ 0.3$^{\star}$ & 1.3 $\pm$ 0.3 \\
10 & 1.6 $\pm$ 0.2         & 1.4 $\pm$ 0.2 \\
20 & 1.5 $\pm$ 0.2         & 1.6 $\pm$ 0.2 \\
30 & 1.5 $\pm$ 0.3         & 1.7 $\pm$ 0.3 \\
40 & 1.6 $\pm$ 0.2         & 1.9 $\pm$ 0.2 \\
\hline

\end{tabular}
\end{table}

%% file: table/task_statistics.tex
\begin{table}[htb]
\centering
\caption{Dataset statistics.}
\begin{tabular}{lccc}
\toprule
\textbf{Dataset} & \textbf{Train} & \textbf{Seen} & \textbf{Unseen} \\
\midrule
ScienceWorld & 1,483 & 194 & 211 \\
ALFWorld & 3,211 & 140 & 134 \\
\bottomrule
\end{tabular}
\label{tab:dataset_statistics}
\end{table}

%% file: box/traj.tex
\begin{tcolorbox}[breakable,center,breakable,title=\Large\centering{Training Dataset Structure}]\label{box:traj}
\textcolor{my_green}{\textbf{High-Level SFT Data:}}\\
\textbf{Input}: \{ \textcolor{my_red}{high prompt}, \textcolor{my_red}{task description}, \textcolor{my_blue}{current observation},  \textcolor{my_purple}{completed subtasks}, \textcolor{my_orange}{previous local progress} \}\\
\textbf{Target}: \textcolor{my_purple}{next subtask}
\\\\
\textcolor{my_green}{\textbf{Low-Level SFT Data:}}\\
\textbf{Input}: \{\textcolor{my_red}{low prompt}, \textcolor{my_red}{subtask}, \textcolor{my_blue}{current observation}, \textcolor{my_orange}{local progress}\} \\ 
\textbf{Target}: \textcolor{my_purple}{action}
\\\\
\textcolor{my_green}{\textbf{Local-Progress SFT Data:}}\\
\textbf{Input}: \{\textcolor{my_red}{local progress prompt}, \textcolor{my_red}{subtask}, \textcolor{my_purple}{executed action}, \textcolor{my_blue}{resulting observation}, \textcolor{my_orange}{previous local progress}\}  \\ 
\textbf{Target}: \textcolor{my_orange}{updated local progress}
\end{tcolorbox}

%% file: box/lp_prompt.tex
\begin{tcolorbox}[
  breakable,
  title=\Large\centering Local Progress Prompt (\textsc{DeepSeek}-ALFWorld),
  before skip=-6pt,
  top=2pt,
  bottom=2pt,
  boxsep=2pt
]

\small
You are an AI agent responsible for updating \emph{local progress}, a short cumulative summary of what has been achieved within the current subtask.

\noindent\textbf{Inputs:}
\begin{itemize}[leftmargin=10pt, itemsep=0pt, topsep=2pt]
  \item current subtask
  \item previous local progress
  \item current action
  \item observation
\end{itemize}

\noindent\textbf{Global Rules:}
\begin{enumerate}\setlength{\itemsep}{0.1em}
  \item \textbf{Exact token matching:}
  All object and location names \textbf{MUST EXACTLY} match strings in the subtask or observation.
  Do \textbf{NOT} rephrase or normalize names.
  
  \item \textbf{No invented facts:}
  Do \textbf{NOT} infer properties, conditions, or failures unless explicitly stated.

  \item \textbf{Task type:}
  Subtasks starting with \texttt{Locate and pick up} or \texttt{Locate and use} are \textbf{Locate} tasks;
  all others are \textbf{Non-Locate} tasks.
\end{enumerate}

\noindent\textbf{Locate Tasks}

\noindent\textbf{Output:}
\texttt{<progress sentence> || [Checked: loc1, loc2, \dots]}

\begin{enumerate}\setlength{\itemsep}{0.1em}
  \item \textbf{Checked update:}
  Retain all previous locations.
  Add the current location \textbf{ONLY IF} a search action is taken and the target is confirmed \textbf{NOT} present.

  \item \textbf{Termination:}
  If the target is found or picked up, the search ends and must not continue.

  \item \textbf{New + Except:}
  For \texttt{new <OBJ> except <OBJ> N}, any different-numbered \texttt{<OBJ>} completes the subtask immediately.

  \item \textbf{Unsuccessful search:}
  The sentence must include \texttt{unchecked} and imply door states (opened / closed),
  without mentioning specific unchecked locations.

  \item \textbf{Language constraints:}
  Do \textbf{NOT} mention checked locations, reuse fixed templates,
  or repeat more than \textbf{3 consecutive words} from the previous progress.
\end{enumerate}

\noindent\textbf{Non-Locate Tasks (Place / Clean / Heat / Cool / Use)}

\noindent\textbf{Output:}
\texttt{<progress sentence>}

\begin{enumerate}\setlength{\itemsep}{0.1em}
  \item Do \textbf{NOT} include \texttt{Checked} or any tags.
  Mention the object \textbf{ONLY IF} it appears in the current action.

  \item \textbf{No existence or location statements:}
  Do \textbf{NOT} state or imply object presence, absence, containment, or discovery.

  \item \textbf{No failure reasoning:}
  Do \textbf{NOT} explain progress via unmet conditions or missing objects.

  \item \textbf{Assumed availability:}
  Treat the target object as available by definition of the subtask.

  \item \textbf{Allowed content only:}
  Describe only the executed operation or its direct state change.
\end{enumerate}

\end{tcolorbox}

%% file: box/prompt.tex
\begin{tcolorbox}[ breakable,
    title=\Large\centering{Prompt in Training and Inference},
  top=2pt,
  bottom=2pt,
  boxsep=2pt
                ]
\textcolor{my_green}{\textbf{High-Level Prompt:}}\\
You are a high-level planner. Based on the state (task description, historical subtasks, last subtask progress and current observation), please generate a clear and simple subtask.
\\\\
\textcolor{my_green}{\textbf{Low-Level Prompt:}}\\
You are a low-level action executor. Based on the current subtask, observation and local progress, please generate a executable action and determine if the subtask is completed (True/False).
\\\\
\textcolor{my_green}{\textbf{Local-Progress Prompt:}}\\
{You are an AI agent responsible for updating local progress within a subtask. Based on the current subtask, the previous local progress, the current action, and the resulting observation, update the local progress.}
\end{tcolorbox}

%% file: table/hyperparameter_one.tex
\begin{table}[t]
\vspace{-2mm}
\setlength{\tabcolsep}{6pt}
\renewcommand{\arraystretch}{1.15}
\caption{STEP-HRL hyperparameters.}
\label{tab:para}
\centering
\begin{tabular}{l c p{0.55\columnwidth}}
\toprule
\textbf{Hyperparameter} & \textbf{Value} \\
\midrule
\multicolumn{2}{l}{\emph{Optimization}} \\
batch size & 128 \\
batch size per device & 8 \\
gradient accumulation steps & 2 \\
optimizer & AdamW \\
actor learning rate & $1\times10^{-5}$ \\
critic learning rate & $1\times10^{-4}$ \\
sft learning rate & $1\times10^{-4}$ \\
\midrule
\multicolumn{2}{l}{\emph{Training schedule}} \\
sft epochs & 5 \\
orl epochs & 3 \\
orl warmup steps & 100 \\
\midrule
\multicolumn{2}{l}{\emph{RL-specific}} \\
discount factor $\gamma$ & 0.99 \\
advantage weighted factor $\beta$ & 1 \\
soft update $\tau_1$ & 0.2 \\
expectile parameter $\tau_2$ & 0.95 \\
\midrule
\multicolumn{2}{l}{\emph{Generation}} \\
sampling temperature & 0.7 \\
max new tokens ($\pi_\theta^h$, $\pi_\theta^l$) & 32 \\
max new tokens ($\pi_\theta^p$) & 150 \\
\midrule
\multicolumn{2}{l}{\emph{LoRA}} \\
lora $r$ & 16 \\
lora alpha & 32 \\
lora dropout & 0.05 \\
lora target modules & \texttt{q\_proj, k\_proj,}\\
& \texttt{v\_proj, o\_proj} \\
\midrule
\multicolumn{2}{l}{\emph{Data}} \\
data mixture ratio & 1:2 \\
env limit steps & 50 \\
\bottomrule
\end{tabular}
\end{table}

%% file: table/eval_alfworld.tex
\begin{table}[!ht]
\caption{Evaluation details on ALFWorld unseen task.}
\label{tab:eval_alfworld}
\centering
\small
\setlength{\tabcolsep}{6pt}
\renewcommand{\arraystretch}{1.15}
\begin{tabular}{c l c c c c}
\toprule
\textbf{Task ID} &
\textbf{Task Name} &
\textbf{\#Variants} &
\textbf{Success Rate (\%)} &
\textbf{Avg. Steps (Succ.)} &
\textbf{Avg. Steps (All)} \\
\midrule
1 & Pick\&Place        & 24 & 100.0 & 12.7 & 12.7 \\
2 & Examine in Light   & 18 & 100.0 & 13.3 & 13.3 \\
3 & Clean\&Place       & 31 & 100.0 & 10.6 & 10.6 \\
4 & Heat\&Place        & 23 & 100.0 & 17.1 & 17.1 \\
5 & Cool\&Place        & 21 & 100.0 & 15.7 & 15.7 \\
6 & Pick Two\&Place    & 17 & 82.4  & 21.8 & 26.6 \\
\midrule
\textbf{Total} & -- & \textbf{134} & \textbf{97.8} & \textbf{14.7} & \textbf{15.3} \\
\bottomrule
\end{tabular}
\end{table}

%% file: table/eval_scienceworld.tex
\begin{table}[!ht]
\caption{Evaluation details on ScienceWorld unseen task.}
\label{tab:eval_scienceworld}
\centering
\small
\setlength{\tabcolsep}{5pt}
\renewcommand{\arraystretch}{1.12}
\begin{tabular}{c l c c c c} 
\toprule
\textbf{Task ID} &
\textbf{Task Name} &
\textbf{\#Variants} &
\textbf{Avg Score} &
\textbf{Avg. Steps (Succ.)} &
\textbf{Avg. Steps (All)} \\
\midrule
0  & boil                                              & 9  & 68.9  & 45.0 & 48.3 \\
1  & change-the-state-of-matter-of                     & 9  & 62.2  & 34.0 & 46.4 \\
2  & chemistry-mix                                     & 8  & 67.8  & 20.8 & 25.5 \\
3  & chemistry-mix-paint-secondary-color               & 9  & 88.9  & 9.5  & 9.4  \\
4  & chemistry-mix-paint-tertiary-color                & 9  & 54.4  & 16.8 & 13.7 \\
5  & find-animal                                       & 10 & 100.0 & 11.6 & 11.6 \\
6  & find-living-thing                                 & 10 & 100.0 & 11.6 & 11.6 \\
7  & find-non-living-thing                             & 10 & 100.0 & 5.8  & 5.8  \\
8  & find-plant                                        & 10 & 97.5  & 10.0 & 14.0 \\
9  & freeze                                            & 9  & 55.0  & 29.0 & 42.6 \\
10 & grow-fruit                                        & 10 & 43.4  & --   & 46.4 \\
11 & grow-plant                                        & 10 & 98.8  & 34.8 & 35.4 \\
12 & identify-life-stages-1                            & 5  & 77.0  & 25.0 & 27.8 \\
13 & identify-life-stages-2                            & 4  & 25.0  & 5.0  & 6.2  \\
17 & lifespan-longest-lived                            & 10 & 100.0 & 4.0  & 4.0  \\
18 & lifespan-longest-lived-then-shortest-lived        & 10 & 100.0 & 5.0  & 5.0  \\
19 & lifespan-shortest-lived                           & 10 & 100.0 & 4.0  & 4.0  \\
20 & measure-melting-point-known-substance             & 10 & 39.1  & --   & 28.5 \\
22 & melt                                              & 9  & 61.9  & 34.7 & 39.8 \\
25 & power-component                                   & 5  & 100.0 & 11.8 & 11.8 \\
26 & power-component-renewable-vs-nonrenewable-energy  & 5  & 21.2  & --   & 37.8 \\
27 & test-conductivity                                 & 10 & 78.3  & 16.4 & 19.7 \\
28 & test-conductivity-of-unknown-substances           & 10 & 70.0  & 17.1 & 18.7 \\
29 & use-thermometer                                   & 10 & 100.0 & 17.6 & 17.6 \\
\midrule
\textbf{Total} & -- & \textbf{211} & \textbf{77.8} & \textbf{15.45} & \textbf{22.0} \\
\bottomrule
\end{tabular}
\end{table}

%% file: box/case1.tex
\begin{figure}[!htp]
\centering
\scalebox{1.}{
\begin{tcolorbox}[center,breakable,title=\Large\centering{Trajectory Comparison}]
\columnseprule=0.5pt

\noindent
\begin{minipage}[t]{0.48\linewidth}
\vspace{0pt} 
\begin{center}
{\Large Normal}
\end{center}

\textcolor{my_green}{\textbf{Task Description:}} 
Your task is to find a(n) non-living thing. First, focus on the thing. Then, move it to the purple box in the bathroom. \\

\textcolor{my_green}{\textbf{Observation:}} 
This room is called the kitchen. In it, you see the agent, a substance called air, a chair, a counter, a cupboard, a freezer, a fridge, a glass jar, a lighter, an oven, a painting, a sink, soap, a stopwatch, a stove, a table with a glass cup, and a thermometer. 
You also see: a door to the bathroom (closed), a door to the hallway (closed), and a door to the outside (closed). \\
\textcolor{my_purple}{\textbf{Action:}} open door to bathroom \\
\textcolor{my_blue}{\textbf{Reward:}} $0.08$ \\
\textcolor{my_green}{\textbf{Observation:}} The door is now open. \\
\textcolor{my_purple}{\textbf{Action:}} go to bathroom \\
\textcolor{my_blue}{\textbf{Reward:}} $0.17$ \\
\textcolor{my_green}{\textbf{Observation:}} 
This room is called the bathroom. In it, you see a bathtub, a drawing, a glass cup, a purple box, a sink, and a toilet. \\
\textcolor{my_purple}{\textbf{Action:}} focus on cup \\
\textcolor{my_blue}{\textbf{Reward:}} $0.50$ \\
\textcolor{my_green}{\textbf{Observation:}} You focus on the glass cup. \\
\textcolor{my_purple}{\textbf{Action:}} move cup to purple box \\
\textcolor{my_blue}{\textbf{Reward:}} $0.25$ \\
\textcolor{my_green}{\textbf{Observation:}} You move the glass cup to the purple box. \\
\textbf{done.}




\end{minipage}
\hfill
\begin{minipage}[t]{0.48\linewidth}
\vspace{0pt} %
\begin{center}
{\Large STEP-HRL}
\end{center}

\textcolor{my_green}{\textbf{Task Description:}} 
Your task is to find a(n) non-living thing. First, focus on the thing. Then, move it to the purple box in the bathroom. \\

\textcolor{my_green}{\textbf{Observation:}} 
This room is called the kitchen. In it, you see the agent, a substance called air, a chair, a counter, a cupboard, a freezer, a fridge, a glass jar, a lighter, an oven, a painting, a sink, soap, a stopwatch, a stove, a table with a glass cup, and a thermometer. 
You also see: a door to the bathroom (closed), a door to the hallway (closed), and a door to the outside (closed). \\
\textcolor{my_red}{\textbf{Subtask 1: Navigate to the bathroom and focus on any non-living object.}} \\
\textcolor{my_purple}{\textbf{Action:}} open door to bathroom \\
\textcolor{my_blue}{\textbf{Reward:}} $0.08$ \\
\textcolor{my_green}{\textbf{Observation:}} 
The door is now open. \\
\textcolor{my_orange}{\textbf{Local Progress:}} 
Agent is in the kitchen and the door to the bathroom is open, not yet in the bathroom. \texttt{[Route: kitchen]} \\
\textcolor{my_purple}{\textbf{Action:}} go to bathroom \\
\textcolor{my_blue}{\textbf{Reward:}} $0.17$ \\
\textcolor{my_green}{\textbf{Observation:}} 
This room is called the bathroom. In it, you see a bathtub, a drawing, a glass cup, a purple box, a sink, and a toilet. \\
\textcolor{my_orange}{\textbf{Local Progress:}} 
Agent is now in the bathroom, ready to search for a non-living thing. 
\texttt{[Route: kitchen $\rightarrow$ bathroom]} \\
\textcolor{my_purple}{\textbf{Action:}} focus on cup \\
\textcolor{my_blue}{\textbf{Reward:}} $0.50$ \\
\textcolor{my_green}{\textbf{Observation:}} 
You focus on the glass cup. \\
\textcolor{my_orange}{\textbf{Local Progress:}} 
Agent has found and focused on a glass cup, a non-living thing, in the bathroom. 
\texttt{[Route: kitchen $\rightarrow$ bathroom]} \\

\textcolor{my_red}{\textbf{Subtask 2: Move the non-living thing to the purple box in the bathroom.}} \\
\textcolor{my_purple}{\textbf{Action:}} move cup to purple box \\
\textcolor{my_blue}{\textbf{Reward:}} $0.25$ \\
\textcolor{my_green}{\textbf{Observation:}} 
You move the glass cup to the purple box. \\
\textbf{done.}
\end{minipage}

\end{tcolorbox}
}
\caption{
Case study comparing a standard ReAct-style agent and STEP-HRL on a ScienceWorld task. Normal paradigms make decisions based on the accumulated interaction trajectory, which tends to collapse in long-horizon complex tasks.
In contrast, STEP-HRL makes decisions based solely on the current observation, augmented with high-level subtask and step-wise local progress, enabling step-level decision making.
}
\label{fig:case1}
\end{figure}

%% file: box/case2.tex
\scalebox{1.}{
\begin{tcolorbox}[center,title=\Large\centering{STEP-HRL in Long-Horizon Complex Task}]
\setlength{\parindent}{0pt}
\setlength{\parskip}{2pt}

\textcolor{my_green}{\textbf{Task Description:}} 
Your task is to boil water. For compounds without a boiling point, combusting the substance is also acceptable. First, focus on the substance. Then, take actions that will cause it to change its state of matter. 

\textcolor{my_green}{\textbf{Observation:}} 
You are in art studio. You also see: A door to the hallway (that is closed). %

\textcolor{my_red}{\textbf{Subtask 1: Navigate to kitchen}}\\
\textcolor{my_purple}{\textbf{Action:}} open door to hallway \;
\textcolor{my_blue}{\textbf{Reward:}} $0.00$ \;
\textcolor{my_green}{\textbf{Observation:}} The door is now open. \;
\textcolor{my_orange}{\textbf{Local Progress:}} Navigating to kitchen and the door to hallway is now open. \;
\textcolor{my_purple}{\textbf{Action:}} go to hallway \;
\textcolor{my_blue}{\textbf{Reward:}} $0.00$ \;
\textcolor{my_green}{\textbf{Observation:}} You move to the hallway. \;
\textcolor{my_orange}{\textbf{Local Progress:}} Moving from art studio to hallway \texttt{[Route: art studio -> hallway]} \;
\textcolor{my_purple}{\textbf{Action:}} open door to kitchen \;
\textcolor{my_blue}{\textbf{Reward:}} $0.00$ \;
\textcolor{my_green}{\textbf{Observation:}} The door is now open. \;
\textcolor{my_orange}{\textbf{Local Progress:}} The kitchen door is open and ready to go. \texttt{[Route: art studio -> hallway]} \;
\textcolor{my_purple}{\textbf{Action:}} go to kitchen \;
\textcolor{my_blue}{\textbf{Reward:}} $0.00$ \;
\textcolor{my_green}{\textbf{Observation:}} You move to the kitchen. \;
\textcolor{my_orange}{\textbf{Local Progress:}} Arrived in the kitchen. \texttt{[Route: art studio -> hallway -> kitchen]}\\

\textcolor{my_red}{\textbf{Subtask 2: Prepare tools for measuring temperature and boiling}}\\
\textcolor{my_purple}{\textbf{Action:}} pick up thermometer \;
\textcolor{my_blue}{\textbf{Reward:}} $0.00$ \;
\textcolor{my_green}{\textbf{Observation:}} You move the thermometer to the inventory. \;
\textcolor{my_orange}{\textbf{Local Progress:}} Thermometer obtained. \;
\textcolor{my_purple}{\textbf{Action:}} open cupboard \;
\textcolor{my_blue}{\textbf{Reward:}} $0.00$ \;
\textcolor{my_green}{\textbf{Observation:}} The cupboard is now open. \;
\textcolor{my_orange}{\textbf{Local Progress:}} Thermometer obtained; cupboard opened, revealing a metal pot and other items. \;
\textcolor{my_purple}{\textbf{Action:}} pick up metal pot \;
\textcolor{my_blue}{\textbf{Reward:}} $0.00$ \;
\textcolor{my_green}{\textbf{Observation:}} You move the metal pot to the inventory. \;
\textcolor{my_orange}{\textbf{Local Progress:}} Metal pot obtained; thermometer already in inventory.\\

\textcolor{my_red}{\textbf{Subtask 3: Fill metal pot with water}}\\
\textcolor{my_purple}{\textbf{Action:}} move metal pot to sink \;
\textcolor{my_blue}{\textbf{Reward:}} $0.00$ \;
\textcolor{my_green}{\textbf{Observation:}} You move the metal pot to the sink. \;
\textcolor{my_orange}{\textbf{Local Progress:}} The metal pot is now in the sink, ready to be filled. \;
\textcolor{my_purple}{\textbf{Action:}} activate sink \;
\textcolor{my_blue}{\textbf{Reward:}} $0.03$ \;
\textcolor{my_green}{\textbf{Observation:}} The sink is now activated. \;
\textcolor{my_orange}{\textbf{Local Progress:}} The metal pot is now filled with water in the sink. \;
\textcolor{my_purple}{\textbf{Action:}} deactivate sink \;
\textcolor{my_blue}{\textbf{Reward:}} $0.00$ \;
\textcolor{my_green}{\textbf{Observation:}} The sink is now deactivated. \;
\textcolor{my_orange}{\textbf{Local Progress:}} The metal pot remains filled with water in the sink, which is now turned off. \;
\textcolor{my_purple}{\textbf{Action:}} pick up metal pot \;
\textcolor{my_blue}{\textbf{Reward:}} $0.00$ \;
\textcolor{my_green}{\textbf{Observation:}} You move the metal pot to the inventory. \;
\textcolor{my_orange}{\textbf{Local Progress:}} The metal pot containing water is now in the agent's inventory.\\

\textcolor{my_red}{\textbf{Subtask 4: Focus on substance}}\\
\textcolor{my_purple}{\textbf{Action:}} focus on substance in metal pot \;
\textcolor{my_blue}{\textbf{Reward:}} $0.67$ \;
\textcolor{my_green}{\textbf{Observation:}} You focus on the water. \;
\textcolor{my_orange}{\textbf{Local Progress:}} You focus on the substance called water in the metal pot.\\

\textcolor{my_red}{\textbf{Subtask 5: Heat water with stove in kitchen}}\\
\textcolor{my_purple}{\textbf{Action:}} move metal pot to stove \;
\textcolor{my_blue}{\textbf{Reward:}} $0.02$ \;
\textcolor{my_green}{\textbf{Observation:}} You move the metal pot to the stove. \;
\textcolor{my_orange}{\textbf{Local Progress:}} Metal pot containing water is now on the stove. \;
\textcolor{my_purple}{\textbf{Action:}} activate stove \;
\textcolor{my_blue}{\textbf{Reward:}} $0.00$ \;
\textcolor{my_green}{\textbf{Observation:}} The stove appears broken, and can't be activated or deactivated. \;
\textcolor{my_orange}{\textbf{Local Progress:}} The stove is broken and cannot be activated; the metal pot with water remains on it.\\

\textcolor{my_red}{\textbf{Subtask 6: Navigate to foundry to heat water because stove is broken}}\\
\textcolor{my_purple}{\textbf{Action:}} pick up metal pot \;
\textcolor{my_blue}{\textbf{Reward:}} $0.00$ \;
\textcolor{my_green}{\textbf{Observation:}} You move the metal pot to the inventory. \;
\textcolor{my_orange}{\textbf{Local Progress:}} Agent has picked up the metal pot containing water. \;
\textcolor{my_purple}{\textbf{Action:}} open door to outside \;
\textcolor{my_blue}{\textbf{Reward:}} $0.00$ \;
\textcolor{my_green}{\textbf{Observation:}} The door is now open. \;
\textcolor{my_orange}{\textbf{Local Progress:}} Agent is still in the kitchen with the pot of water, and the door to outside is open. \texttt{[Route: kitchen]} \;
\textcolor{my_purple}{\textbf{Action:}} go to outside \;
\textcolor{my_blue}{\textbf{Reward:}} $0.00$ \;
\textcolor{my_green}{\textbf{Observation:}} You move to the outside. \;
\textcolor{my_orange}{\textbf{Local Progress:}} Agent is now outside, heading toward the foundry. \texttt{[Route: kitchen -> outside]} \;
\textcolor{my_purple}{\textbf{Action:}} open door

\end{tcolorbox}
}

{\centering
\scalebox{1.}{
\begin{tcolorbox}[center]
\setlength{\parindent}{0pt}
\setlength{\parskip}{2pt}

to foundry \;
\textcolor{my_blue}{\textbf{Reward:}} $0.00$ \;
\textcolor{my_green}{\textbf{Observation:}} The door is now open. \;
\textcolor{my_orange}{\textbf{Local Progress:}} Agent has opened the foundry door but has not yet entered. \texttt{[Route: kitchen -> outside]} \;
\textcolor{my_purple}{\textbf{Action:}} go to foundry \;
\textcolor{my_blue}{\textbf{Reward:}} $0.00$ \;
\textcolor{my_green}{\textbf{Observation:}} You move to the foundry. \;
\textcolor{my_orange}{\textbf{Local Progress:}} Agent has entered the foundry. \texttt{[Route: kitchen -> outside -> foundry]}\\

\textcolor{my_red}{\textbf{Subtask 7: Heat water with blast furnace}}\\
\textcolor{my_purple}{\textbf{Action:}} open blast furnace \;
\textcolor{my_blue}{\textbf{Reward:}} $0.00$ \;
\textcolor{my_green}{\textbf{Observation:}} The blast furnace is now open. \;
\textcolor{my_orange}{\textbf{Local Progress:}} The blast furnace is open and ready to be used. \;
\textcolor{my_purple}{\textbf{Action:}} move metal pot to blast furnace \;
\textcolor{my_blue}{\textbf{Reward:}} $0.01$ \;
\textcolor{my_green}{\textbf{Observation:}} You move the metal pot to the blast furnace. \;
\textcolor{my_orange}{\textbf{Local Progress:}} The metal pot containing water is now inside the open blast furnace. \;
\textcolor{my_purple}{\textbf{Action:}} activate blast furnace \;
\textcolor{my_blue}{\textbf{Reward:}} $0.02$ \;
\textcolor{my_green}{\textbf{Observation:}} The blast furnace is now activated. \;
\textcolor{my_orange}{\textbf{Local Progress:}} The blast furnace is activated and heating the metal pot containing water.\\

\textcolor{my_red}{\textbf{Subtask 8: Monitor water temperature}}\\
\textcolor{my_purple}{\textbf{Action:}} examine substance in metal pot \;
\textcolor{my_blue}{\textbf{Reward:}} $0.02$ \;
\textcolor{my_green}{\textbf{Observation:}} a substance called water \;
\textcolor{my_orange}{\textbf{Local Progress:}} Examined water in the pot; temperature not yet measured. \;
\textcolor{my_purple}{\textbf{Action:}} use thermometer in inventory on substance in metal pot \;
\textcolor{my_blue}{\textbf{Reward:}} $0.00$ \;
\textcolor{my_green}{\textbf{Observation:}} the thermometer measures a temperature of 50 degrees celsius \;
\textcolor{my_orange}{\textbf{Local Progress:}} Water temperature measured at 50\textdegree C. \;
\textcolor{my_purple}{\textbf{Action:}} examine substance in metal pot \;
\textcolor{my_blue}{\textbf{Reward:}} $0.00$ \;
\textcolor{my_green}{\textbf{Observation:}} a substance called water \;
\textcolor{my_orange}{\textbf{Local Progress:}} Water temperature measured at 50\textdegree C. \;
\textcolor{my_purple}{\textbf{Action:}} wait1 \;
\textcolor{my_blue}{\textbf{Reward:}} $0.23$ \;
\textcolor{my_green}{\textbf{Observation:}} You decide to wait for 1 iterations.
\par\textbf{done.}

\end{tcolorbox}
}
\par}

%% file: custom.bib
@inproceedings{yao2022react,
  title={React: Synergizing reasoning and acting in language models},
  author={Yao, Shunyu and Zhao, Jeffrey and Yu, Dian and Du, Nan and Shafran, Izhak and Narasimhan, Karthik R and Cao, Yuan},
  booktitle={The eleventh international conference on learning representations},
  year={2022}
}

@article{shinn2023reflexion,
  title={Reflexion: Language agents with verbal reinforcement learning},
  author={Shinn, Noah and Cassano, Federico and Gopinath, Ashwin and Narasimhan, Karthik and Yao, Shunyu},
  journal={Advances in Neural Information Processing Systems},
  volume={36},
  pages={8634--8652},
  year={2023}
}

@article{wangvoyager,
  title={Voyager: An Open-Ended Embodied Agent with Large Language Models},
  author={Wang, Guanzhi and Xie, Yuqi and Jiang, Yunfan and Mandlekar, Ajay and Xiao, Chaowei and Zhu, Yuke and Fan, Linxi and Anandkumar, Anima},
  journal={Transactions on Machine Learning Research},
  year={2023}
}

@article{li2022pre,
  title={Pre-trained language models for interactive decision-making},
  author={Li, Shuang and Puig, Xavier and Paxton, Chris and Du, Yilun and Wang, Clinton and Fan, Linxi and Chen, Tao and Huang, De-An and Aky{\"u}rek, Ekin and Anandkumar, Anima and others},
  journal={Advances in Neural Information Processing Systems},
  volume={35},
  pages={31199--31212},
  year={2022}
}

@inproceedings{zeng-etal-2024-agenttuning,
    title = "{A}gent{T}uning: Enabling Generalized Agent Abilities for {LLM}s",
    author = "Zeng, Aohan  and
      Liu, Mingdao  and
      Lu, Rui  and
      Wang, Bowen  and
      Liu, Xiao  and
      Dong, Yuxiao  and
      Tang, Jie",
    editor = "Ku, Lun-Wei  and
      Martins, Andre  and
      Srikumar, Vivek",
    booktitle = "Findings of the Association for Computational Linguistics: ACL 2024",
    month = aug,
    year = "2024",
    address = "Bangkok, Thailand",
    publisher = "Association for Computational Linguistics",
    url = "https://aclanthology.org/2024.findings-acl.181/",
    doi = "10.18653/v1/2024.findings-acl.181",
    pages = "3053--3077"
}

@article{lin2023swiftsage,
  title={Swiftsage: A generative agent with fast and slow thinking for complex interactive tasks},
  author={Lin, Bill Yuchen and Fu, Yicheng and Yang, Karina and Brahman, Faeze and Huang, Shiyu and Bhagavatula, Chandra and Ammanabrolu, Prithviraj and Choi, Yejin and Ren, Xiang},
  journal={Advances in Neural Information Processing Systems},
  volume={36},
  pages={23813--23825},
  year={2023}
}

@inproceedings{zhao-etal-2024-epo,
    title = "{EPO}: Hierarchical {LLM} Agents with Environment Preference Optimization",
    author = "Zhao, Qi  and
      Fu, Haotian  and
      Sun, Chen  and
      Konidaris, George",
    editor = "Al-Onaizan, Yaser  and
      Bansal, Mohit  and
      Chen, Yun-Nung",
    booktitle = "Proceedings of the 2024 Conference on Empirical Methods in Natural Language Processing",
    month = nov,
    year = "2024",
    address = "Miami, Florida, USA",
    publisher = "Association for Computational Linguistics",
    url = "https://aclanthology.org/2024.emnlp-main.367/",
    doi = "10.18653/v1/2024.emnlp-main.367",
    pages = "6401--6415"
}

@inproceedings{song-etal-2024-eto,
    title = "Trial and Error: Exploration-Based Trajectory Optimization of {LLM} Agents",
    author = "Song, Yifan  and
      Yin, Da  and
      Yue, Xiang  and
      Huang, Jie  and
      Li, Sujian  and
      Lin, Bill Yuchen",
    editor = "Ku, Lun-Wei  and
      Martins, Andre  and
      Srikumar, Vivek",
    booktitle = "Proceedings of the 62nd Annual Meeting of the Association for Computational Linguistics (Volume 1: Long Papers)",
    month = aug,
    year = "2024",
    address = "Bangkok, Thailand",
    publisher = "Association for Computational Linguistics",
    url = "https://aclanthology.org/2024.acl-long.409/",
    doi = "10.18653/v1/2024.acl-long.409",
    pages = "7584--7600"
}

@inproceedings{xu2024strategicAgents,
  title={Language agents with reinforcement learning for strategic play in the Werewolf game},
  author={Xu, Zelai and Yu, Chao and Fang, Fei and Wang, Yu and Wu, Yi},
  booktitle={Proceedings of the 41st International Conference on Machine Learning},
  pages={55434--55464},
  year={2024}
}

@article{pang2024kalm,
  title={Kalm: Knowledgeable agents by offline reinforcement learning from large language model rollouts},
  author={Pang, Jing-Cheng and Yang, Si-Hang and Li, Kaiyuan and Zhang, Jiaji and Chen, Xiong-Hui and Tang, Nan and Yu, Yang},
  journal={Advances in Neural Information Processing Systems},
  volume={37},
  pages={126620--126652},
  year={2024}
}

@article{peiyuan2024agile,
  title={Agile: A novel reinforcement learning framework of llm agents},
  author={Peiyuan, Feng and He, Yichen and Huang, Guanhua and Lin, Yuan and Zhang, Hanchong and Zhang, Yuchen and Li, Hang},
  journal={Advances in Neural Information Processing Systems},
  volume={37},
  pages={5244--5284},
  year={2024}
}

@article{chen2021decision,
  title={Decision transformer: Reinforcement learning via sequence modeling},
  author={Chen, Lili and Lu, Kevin and Rajeswaran, Aravind and Lee, Kimin and Grover, Aditya and Laskin, Misha and Abbeel, Pieter and Srinivas, Aravind and Mordatch, Igor},
  journal={Advances in neural information processing systems},
  volume={34},
  pages={15084--15097},
  year={2021}
}

@article{janner2021offline,
  title={Offline reinforcement learning as one big sequence modeling problem},
  author={Janner, Michael and Li, Qiyang and Levine, Sergey},
  journal={Advances in neural information processing systems},
  volume={34},
  pages={1273--1286},
  year={2021}
}

@article{ni2023transformers,
  title={When do transformers shine in rl? decoupling memory from credit assignment},
  author={Ni, Tianwei and Ma, Michel and Eysenbach, Benjamin and Bacon, Pierre-Luc},
  journal={Advances in Neural Information Processing Systems},
  volume={36},
  pages={50429--50452},
  year={2023}
}

@article{vaswani2017attention,
  title={Attention is all you need},
  author={Vaswani, Ashish and Shazeer, Noam and Parmar, Niki and Uszkoreit, Jakob and Jones, Llion and Gomez, Aidan N and Kaiser, {\L}ukasz and Polosukhin, Illia},
  journal={Advances in neural information processing systems},
  volume={30},
  year={2017}
}

@article{zhou2025mem1,
  title={MEM1: Learning to Synergize Memory and Reasoning for Efficient Long-Horizon Agents},
  author={Zhou, Zijian and Qu, Ao and Wu, Zhaoxuan and Kim, Sunghwan and Prakash, Alok and Rus, Daniela and Zhao, Jinhua and Low, Bryan Kian Hsiang and Liang, Paul Pu},
  journal={arXiv preprint arXiv:2506.15841},
  year={2025}
}

@article{cherepanov2023recurrent,
  title={Recurrent action transformer with memory},
  author={Cherepanov, Egor and Staroverov, Alexey and Yudin, Dmitry and Kovalev, Alexey K and Panov, Aleksandr I},
  journal={arXiv preprint arXiv:2306.09459},
  year={2023}
}

@article{luo2024efficient,
  title={Efficient recurrent off-policy RL requires a context-encoder-specific learning rate},
  author={Luo, Fan-Ming and Tu, Zuolin and Huang, Zefang and Yu, Yang},
  journal={Advances in Neural Information Processing Systems},
  volume={37},
  pages={48484--48518},
  year={2024}
}

@article{liu2025agentic,
  title={Agentic Reinforcement Learning with Implicit Step Rewards},
  author={Liu, Xiaoqian and Wang, Ke and Wu, Yuchuan and Huang, Fei and Li, Yongbin and Zhang, Junge and Jiao, Jianbin},
  journal={arXiv preprint arXiv:2509.19199},
  year={2025}
}

@inproceedings{zhai2025enhancing,
  title={Enhancing decision-making for llm agents via step-level q-value models},
  author={Zhai, Yuanzhao and Yang, Tingkai and Xu, Kele and Feng, Dawei and Yang, Cheng and Ding, Bo and Wang, Huaimin},
  booktitle={Proceedings of the AAAI Conference on Artificial Intelligence},
  volume={39},
  number={25},
  pages={27161--27169},
  year={2025}
}

@inproceedings{xiong-etal-2024-watch,
    title = "Watch Every Step! {LLM} Agent Learning via Iterative Step-level Process Refinement",
    author = "Xiong, Weimin  and
      Song, Yifan  and
      Zhao, Xiutian  and
      Wu, Wenhao  and
      Wang, Xun  and
      Wang, Ke  and
      Li, Cheng  and
      Peng, Wei  and
      Li, Sujian",
    editor = "Al-Onaizan, Yaser  and
      Bansal, Mohit  and
      Chen, Yun-Nung",
    booktitle = "Proceedings of the 2024 Conference on Empirical Methods in Natural Language Processing",
    month = nov,
    year = "2024",
    address = "Miami, Florida, USA",
    publisher = "Association for Computational Linguistics",
    url = "https://aclanthology.org/2024.emnlp-main.93/",
    doi = "10.18653/v1/2024.emnlp-main.93",
    pages = "1556--1572"
}

@article{hu2025divide,
  title={Divide and Conquer: Grounding LLMs as Efficient Decision-Making Agents via Offline Hierarchical Reinforcement Learning},
  author={Hu, Zican and Liu, Wei and Qu, Xiaoye and Yue, Xiangyu and Chen, Chunlin and Wang, Zhi and Cheng, Yu},
  journal={arXiv preprint arXiv:2505.19761},
  year={2025}
}

@article{rafailov2023dpo,
  title={Direct preference optimization: Your language model is secretly a reward model},
  author={Rafailov, Rafael and Sharma, Archit and Mitchell, Eric and Manning, Christopher D and Ermon, Stefano and Finn, Chelsea},
  journal={Advances in neural information processing systems},
  volume={36},
  pages={53728--53741},
  year={2023}
}

@inproceedings{snellILQL,
  title={Offline RL for Natural Language Generation with Implicit Language Q Learning},
  author={Snell, Charlie Victor and Kostrikov, Ilya and Su, Yi and Yang, Sherry and Levine, Sergey},
  booktitle={The Eleventh International Conference on Learning Representations},
  year={2023}
}

@article{kostrikov2021IQL,
  title={Offline reinforcement learning with implicit q-learning},
  author={Kostrikov, Ilya and Nair, Ashvin and Levine, Sergey},
  journal={arXiv preprint arXiv:2110.06169},
  year={2021}
}

@inproceedings{haarnoja2018soft,
  title={Soft actor-critic: Off-policy maximum entropy deep reinforcement learning with a stochastic actor},
  author={Haarnoja, Tuomas and Zhou, Aurick and Abbeel, Pieter and Levine, Sergey},
  booktitle={International conference on machine learning},
  pages={1861--1870},
  year={2018},
  organization={Pmlr}
}

@article{peng2019AWR,
  title={Advantage-weighted regression: Simple and scalable off-policy reinforcement learning},
  author={Peng, Xue Bin and Kumar, Aviral and Zhang, Grace and Levine, Sergey},
  journal={arXiv preprint arXiv:1910.00177},
  year={2019}
}

@article{nair2020awac,
  title={Awac: Accelerating online reinforcement learning with offline datasets},
  author={Nair, Ashvin and Gupta, Abhishek and Dalal, Murtaza and Levine, Sergey},
  journal={arXiv preprint arXiv:2006.09359},
  year={2020}
}

@article{wang2022scienceworld,
  title={Scienceworld: Is your agent smarter than a 5th grader?},
  author={Wang, Ruoyao and Jansen, Peter and C{\^o}t{\'e}, Marc-Alexandre and Ammanabrolu, Prithviraj},
  journal={arXiv preprint arXiv:2203.07540},
  year={2022}
}

@article{shridhar2020alfworld,
  title={Alfworld: Aligning text and embodied environments for interactive learning},
  author={Shridhar, Mohit and Yuan, Xingdi and C{\^o}t{\'e}, Marc-Alexandre and Bisk, Yonatan and Trischler, Adam and Hausknecht, Matthew},
  journal={arXiv preprint arXiv:2010.03768},
  year={2020}
}

@article{Jiang2023Mistral7,
  title={Mistral 7B},
  author={Albert Qiaochu Jiang and Alexandre Sablayrolles and Arthur Mensch and Chris Bamford and Devendra Singh Chaplot and Diego de Las Casas and Florian Bressand and Gianna Lengyel and Guillaume Lample and Lucile Saulnier and L{\'e}lio Renard Lavaud and Marie-Anne Lachaux and Pierre Stock and Teven Le Scao and Thibaut Lavril and Thomas Wang and Timoth{\'e}e Lacroix and William El Sayed},
  journal={ArXiv},
  year={2023},
  volume={abs/2310.06825},
  url={https://api.semanticscholar.org/CorpusID:263830494}
}

@article{team2024gemma,
  title={Gemma: Open models based on gemini research and technology},
  author={Team, Gemma and Mesnard, Thomas and Hardin, Cassidy and Dadashi, Robert and Bhupatiraju, Surya and Pathak, Shreya and Sifre, Laurent and Rivi{\`e}re, Morgane and Kale, Mihir Sanjay and Love, Juliette and others},
  journal={arXiv preprint arXiv:2403.08295},
  year={2024}
}

@misc{meta2024llama3,
  author       = {{Meta AI}},
  title        = {Introducing Meta LLaMA 3: The Most Capable Openly Available LLM to Date},
  year         = {2024},
  howpublished = {\url{https://ai.meta.com/blog/meta-llama-3/}},
  note         = {Accessed: 2024-03}
}

@article{qiao2024WKM,
  title={Agent planning with world knowledge model},
  author={Qiao, Shuofei and Fang, Runnan and Zhang, Ningyu and Zhu, Yuqi and Chen, Xiang and Deng, Shumin and Jiang, Yong and Xie, Pengjun and Huang, Fei and Chen, Huajun},
  journal={Advances in Neural Information Processing Systems},
  volume={37},
  pages={114843--114871},
  year={2024}
}

@article{hu2022lora,
  title={Lora: Low-rank adaptation of large language models.},
  author={Hu, Edward J and Shen, Yelong and Wallis, Phillip and Allen-Zhu, Zeyuan and Li, Yuanzhi and Wang, Shean and Wang, Lu and Chen, Weizhu and others},
  journal={ICLR},
  volume={1},
  number={2},
  pages={3},
  year={2022}
}

@article{loshchilov2017decoupled,
  title={Decoupled weight decay regularization},
  author={Loshchilov, Ilya and Hutter, Frank},
  journal={arXiv preprint arXiv:1711.05101},
  year={2017}
}

@article{wang2024survey,
  title={A survey on large language model based autonomous agents},
  author={Wang, Lei and Ma, Chen and Feng, Xueyang and Zhang, Zeyu and Yang, Hao and Zhang, Jingsen and Chen, Zhiyuan and Tang, Jiakai and Chen, Xu and Lin, Yankai and others},
  journal={Frontiers of Computer Science},
  volume={18},
  number={6},
  pages={186345},
  year={2024},
  publisher={Springer}
}

@article{guo2024large,
  title={Large language model based multi-agents: A survey of progress and challenges},
  author={Guo, Taicheng and Chen, Xiuying and Wang, Yaqi and Chang, Ruidi and Pei, Shichao and Chawla, Nitesh V and Wiest, Olaf and Zhang, Xiangliang},
  journal={arXiv preprint arXiv:2402.01680},
  year={2024}
}

@article{wei2022chain,
  title={Chain-of-thought prompting elicits reasoning in large language models},
  author={Wei, Jason and Wang, Xuezhi and Schuurmans, Dale and Bosma, Maarten and Xia, Fei and Chi, Ed and Le, Quoc V and Zhou, Denny and others},
  journal={Advances in neural information processing systems},
  volume={35},
  pages={24824--24837},
  year={2022}
}

@article{yao2023tree,
  title={Tree of thoughts: Deliberate problem solving with large language models},
  author={Yao, Shunyu and Yu, Dian and Zhao, Jeffrey and Shafran, Izhak and Griffiths, Tom and Cao, Yuan and Narasimhan, Karthik},
  journal={Advances in neural information processing systems},
  volume={36},
  pages={11809--11822},
  year={2023}
}

@article{zhang2025survey,
  title={A survey on the memory mechanism of large language model-based agents},
  author={Zhang, Zeyu and Dai, Quanyu and Bo, Xiaohe and Ma, Chen and Li, Rui and Chen, Xu and Zhu, Jieming and Dong, Zhenhua and Wen, Ji-Rong},
  journal={ACM Transactions on Information Systems},
  volume={43},
  number={6},
  pages={1--47},
  year={2025},
  publisher={ACM New York, NY}
}

@article{schick2023toolformer,
  title={Toolformer: Language models can teach themselves to use tools},
  author={Schick, Timo and Dwivedi-Yu, Jane and Dess{\`\i}, Roberto and Raileanu, Roberta and Lomeli, Maria and Hambro, Eric and Zettlemoyer, Luke and Cancedda, Nicola and Scialom, Thomas},
  journal={Advances in Neural Information Processing Systems},
  volume={36},
  pages={68539--68551},
  year={2023}
}

@article{qin2023toolllm,
  title={Toolllm: Facilitating large language models to master 16000+ real-world apis},
  author={Qin, Yujia and Liang, Shihao and Ye, Yining and Zhu, Kunlun and Yan, Lan and Lu, Yaxi and Lin, Yankai and Cong, Xin and Tang, Xiangru and Qian, Bill and others},
  journal={arXiv preprint arXiv:2307.16789},
  year={2023}
}

@article{wu2024avatar,
  title={Avatar: Optimizing llm agents for tool usage via contrastive reasoning},
  author={Wu, Shirley and Zhao, Shiyu and Huang, Qian and Huang, Kexin and Yasunaga, Michihiro and Cao, Kaidi and Ioannidis, Vassilis and Subbian, Karthik and Leskovec, Jure and Zou, James Y},
  journal={Advances in Neural Information Processing Systems},
  volume={37},
  pages={25981--26010},
  year={2024}
}

@article{wang2022self,
  title={Self-consistency improves chain of thought reasoning in language models},
  author={Wang, Xuezhi and Wei, Jason and Schuurmans, Dale and Le, Quoc and Chi, Ed and Narang, Sharan and Chowdhery, Aakanksha and Zhou, Denny},
  journal={arXiv preprint arXiv:2203.11171},
  year={2022}
}

@inproceedings{ren2025r2dqg,
  title={R2DQG: a quality meets diversity framework for question generation over knowledge bases},
  author={Ren, Yimeng and Yu, Yanhua and Liao, Lizi and Shang, Yuhu and Lu, Kangkang and Yan, Mingliang},
  booktitle={Proceedings of the Thirty-Fourth International Joint Conference on Artificial Intelligence},
  pages={8231--8240},
  year={2025}
}

@inproceedings{li2025meco,
    title = "Adaptive Tool Use in Large Language Models with Meta-Cognition Trigger",
    author = "Li, Wenjun  and
      Li, Dexun  and
      Dong, Kuicai  and
      Zhang, Cong  and
      Zhang, Hao  and
      Liu, Weiwen  and
      Wang, Yasheng  and
      Tang, Ruiming  and
      Liu, Yong",
    editor = "Che, Wanxiang  and
      Nabende, Joyce  and
      Shutova, Ekaterina  and
      Pilehvar, Mohammad Taher",
    booktitle = "Proceedings of the 63rd Annual Meeting of the Association for Computational Linguistics (Volume 1: Long Papers)",
    month = jul,
    year = "2025",
    address = "Vienna, Austria",
    publisher = "Association for Computational Linguistics",
    url = "https://aclanthology.org/2025.acl-long.655/",
    doi = "10.18653/v1/2025.acl-long.655",
    pages = "13346--13370",
    ISBN = "979-8-89176-251-0"
}

@article{sarch2024vlm,
  title={Vlm agents generate their own memories: Distilling experience into embodied programs of thought},
  author={Sarch, Gabriel and Jang, Lawrence and Tarr, Michael and Cohen, William W and Marino, Kenneth and Fragkiadaki, Katerina},
  journal={Advances in Neural Information Processing Systems},
  volume={37},
  pages={75942--75985},
  year={2024}
}

@article{xu2025mem,
  title={A-mem: Agentic memory for llm agents},
  author={Xu, Wujiang and Liang, Zujie and Mei, Kai and Gao, Hang and Tan, Juntao and Zhang, Yongfeng},
  journal={arXiv preprint arXiv:2502.12110},
  year={2025}
}

@inproceedings{chen2024agentverse,
  title={AgentVerse: Facilitating Multi-Agent Collaboration and Exploring Emergent Behaviors.},
  author={Chen, Weize and Su, Yusheng and Zuo, Jingwei and Yang, Cheng and Yuan, Chenfei and Chan, Chi-Min and Yu, Heyang and Lu, Yaxi and Hung, Yi-Hsin and Qian, Chen and others},
  booktitle={ICLR},
  year={2024}
}

@article{estornell2024acc,
  title={ACC-collab: An actor-critic approach to multi-agent LLM collaboration},
  author={Estornell, Andrew and Ton, Jean-Fran{\c{c}}ois and Yao, Yuanshun and Liu, Yang},
  journal={arXiv preprint arXiv:2411.00053},
  year={2024}
}

@article{bo2024reflective,
  title={Reflective multi-agent collaboration based on large language models},
  author={Bo, Xiaohe and Zhang, Zeyu and Dai, Quanyu and Feng, Xueyang and Wang, Lei and Li, Rui and Chen, Xu and Wen, Ji-Rong},
  journal={Advances in Neural Information Processing Systems},
  volume={37},
  pages={138595--138631},
  year={2024}
}

@inproceedings{zeng2024agenttuning,
  title={Agenttuning: Enabling generalized agent abilities for llms},
  author={Zeng, Aohan and Liu, Mingdao and Lu, Rui and Wang, Bowen and Liu, Xiao and Dong, Yuxiao and Tang, Jie},
  booktitle={Findings of the Association for Computational Linguistics: ACL 2024},
  pages={3053--3077},
  year={2024}
}

@article{chen2023fireact,
  title={Fireact: Toward language agent fine-tuning},
  author={Chen, Baian and Shu, Chang and Shareghi, Ehsan and Collier, Nigel and Narasimhan, Karthik and Yao, Shunyu},
  journal={arXiv preprint arXiv:2310.05915},
  year={2023}
}

@inproceedings{yin2023lumos,
  title={Lumos: Learning agents with unified data, modular design, and open-source llms},
  author={Yin, Da and Brahman, Faeze and Ravichander, Abhilasha and Chandu, Khyathi and Chang, Kai-Wei and Choi, Yejin and Lin, Bill Yuchen},
  booktitle={ICLR 2024 Workshop on Large Language Model (LLM) Agents},
  year={2023}
}

@article{chen2024agent,
  title={Agent-flan: Designing data and methods of effective agent tuning for large language models},
  author={Chen, Zehui and Liu, Kuikun and Wang, Qiuchen and Zhang, Wenwei and Liu, Jiangning and Lin, Dahua and Chen, Kai and Zhao, Feng},
  journal={arXiv preprint arXiv:2403.12881},
  year={2024}
}

@article{ouyang2022training,
  title={Training language models to follow instructions with human feedback},
  author={Ouyang, Long and Wu, Jeffrey and Jiang, Xu and Almeida, Diogo and Wainwright, Carroll and Mishkin, Pamela and Zhang, Chong and Agarwal, Sandhini and Slama, Katarina and Ray, Alex and others},
  journal={Advances in neural information processing systems},
  volume={35},
  pages={27730--27744},
  year={2022}
}

@article{feng2025group,
  title={Group-in-group policy optimization for llm agent training},
  author={Feng, Lang and Xue, Zhenghai and Liu, Tingcong and An, Bo},
  journal={arXiv preprint arXiv:2505.10978},
  year={2025}
}

@article{wen2024reinforcing,
  title={Reinforcing llm agents via policy optimization with action decomposition},
  author={Wen, Muning and Wan, Ziyu and Wang, Jun and Zhang, Weinan and Wen, Ying},
  journal={Advances in Neural Information Processing Systems},
  volume={37},
  pages={103774--103805},
  year={2024}
}

@article{zhai2024fine,
  title={Fine-tuning large vision-language models as decision-making agents via reinforcement learning},
  author={Zhai, Simon and Bai, Hao and Lin, Zipeng and Pan, Jiayi and Tong, Peter and Zhou, Yifei and Suhr, Alane and Xie, Saining and LeCun, Yann and Ma, Yi and others},
  journal={Advances in neural information processing systems},
  volume={37},
  pages={110935--110971},
  year={2024}
}

@inproceedings{szot2023large,
  title={Large language models as generalizable policies for embodied tasks},
  author={Szot, Andrew and Schwarzer, Max and Agrawal, Harsh and Mazoure, Bogdan and Metcalf, Rin and Talbott, Walter and Mackraz, Natalie and Hjelm, R Devon and Toshev, Alexander T},
  booktitle={The Twelfth International Conference on Learning Representations},
  year={2023}
}

@article{shao2024deepseekmath,
  title={Deepseekmath: Pushing the limits of mathematical reasoning in open language models},
  author={Shao, Zhihong and Wang, Peiyi and Zhu, Qihao and Xu, Runxin and Song, Junxiao and Bi, Xiao and Zhang, Haowei and Zhang, Mingchuan and Li, YK and Wu, Yang and others},
  journal={arXiv preprint arXiv:2402.03300},
  year={2024}
}

@article{schulman2017proximal,
  title={Proximal policy optimization algorithms},
  author={Schulman, John and Wolski, Filip and Dhariwal, Prafulla and Radford, Alec and Klimov, Oleg},
  journal={arXiv preprint arXiv:1707.06347},
  year={2017}
}
